\documentclass[fleqn,10pt]{wlscirep}
\usepackage[utf8]{inputenc}
\usepackage[T1]{fontenc}
\usepackage{amsmath}
\usepackage{cite}
\usepackage{wrapfig}
\usepackage{multirow}
\usepackage{graphics} 
\usepackage{epsfig} 
\usepackage{graphicx}
\usepackage{caption}
\usepackage{subcaption}
\usepackage{multirow}
\usepackage{colortbl}

\graphicspath{{./figures/}}

\title{Multi-Stain Multi-Level Convolutional Network for Multi-Tissue Breast Cancer Image Segmentation
}

\author[1]{Akash Modi}
\author[1]{Sumit Kumar Jha}
\author[1,*]{Purnendu Mishra}
\author[1,]{Rajiv Kumar}
\author[1]{Kiran Aatre}
\author[1]{Gursewak Singh}
\author[1]{Shubham Mathur}
\affil[1]{Applied Materials India Pvt. Ltd., Bangalore, India}

\affil[*]{purnendu\_mishra@amat.com (corresponding author)}

\begin{abstract}
Digital pathology and microscopy image analysis are widely employed in the segmentation of digitally scanned IHC slides, primarily to identify cancer and pinpoint regions of interest (ROI) indicative of tumor presence. However, current ROI segmentation models are either stain-specific or suffer from the issues of stain and scanner variance due to different staining protocols or modalities across multiple labs. Also, tissues like Ductal Carcinoma in Situ (DCIS), acini, etc. are often classified as Tumors due to their structural similarities and color compositions. In this paper, we proposed a novel convolutional neural network (CNN) based Multi-class Tissue Segmentation model for histopathology whole-slide Breast slides which classify tumors and segments other tissue regions such as Ducts, acini, DCIS, Squamous epithelium, Blood Vessels, Necrosis, etc. as a separate class. Our unique pixel-aligned non-linear merge across spatial resolutions empowers models with both local and global fields of view for accurate detection of various classes. Our proposed model is also able to separate bad regions such as folds, artifacts, blurry regions, bubbles, etc. from tissue regions using multi-level context from different resolutions of WSI. Multi-phase iterative training with context-aware augmentation and increasing noise was used to efficiently train a multi-stain generic model with partial and noisy annotations from 513 slides. Our training pipeline used 12 million patches generated using context-aware augmentations which made our model stain \& scanner invariant across data sources. To extrapolate stain \& scanner invariance, our model was evaluated on 23000 patches which were for a completely new stain (Hematoxylin and Eosin) from a completely new scanner (Motic) from a different lab. The mean IOU was 0.72 which is on par with model performance on other data sources and scanners. Our results illustrate that the proposed model exhibits remarkable accuracy with consistent performance and broad utility, especially in facilitating pathologists’ assessment of breast cancer.
\end{abstract}
\begin{document}

\flushbottom
\maketitle
%
%
\thispagestyle{empty}


\section*{Introduction}
Cancer is one of the leading healthcare problems worldwide with the second highest cause of death in the US~\cite{diseasecancercollaboration2019global}. According to Cancer Statistics 2021~\cite{siegel2021cancer}, the estimated number of new cancer cases is 1.8 million in the US alone which is equivalent to 5200 cases per day. In women, the most diagnosed cancer type is breast cancer constituting 30\% of the total cases in women. According to the American Cancer Society journal~\cite{siegel2021cancer}, the estimated death of women due to Invasive Breast cancer is approximately 44,000 in the US. The most common type of malignant neoplasm in women is mammary carcinoma increasing at a rate of 0.5\% every year. However, the mortality rate has substantially reduced majorly due to improved imaging techniques such as mammography, sonography, MRI, Magnetic Elastography, Optical imaging, Breast Microwave Imaging, etc. Details of these imaging techniques can be referred to in~\cite{iranmakani2020review}. These advanced imaging techniques lead to early detection which when combined with improved methods of treatment leads to better diagnosis.

Breast carcinoma mainly arises from the terminal duct lobular unit (TDLU). In general, breast cancer is divided into IDC (Invasive ductal carcinoma) and DCIS (ductal carcinoma in situ) based on the infiltration of the neoplasms~\cite{makki2015diversity}. DCIS are non-invasive proliferation of epithelial cells that are confined within ducts and lobules whereas IDC are malignant neoplastic cells that have infiltrated the stroma. Based on the origin of the tumor, it can also be categorized into ductal carcinoma (originating from ducts) and lobular carcinoma (originating from lobules).

Breast cancer detection involves several steps, including the use of various techniques such as Mammograms, diagnostic mammography, and ultrasound to identify abnormalities. However, the definitive diagnosis of breast cancer relies on biopsy. The core needle biopsy (CNB)~\cite{doberneck1980breast} is the predominant method employed today, wherein imaging techniques extract a tissue sample from the identified abnormal region through imaging techniques. Pathologists then analyze the extracted tissue for histological features under a microscope. Unfortunately, the raw, colorless tissue lacks distinct structures like nuclei, cytoplasm, and membrane, necessitating a mandatory staining step ~\cite{alturkistani2015histological} before microscopic analysis. The commonly used H\&E staining imparts a bluish color to nuclei and a pinkish color to the cytoplasm.

These stained tissues are analyzed under the microscope for cell morphological features like texture, size, shape, organization, interactions, spatial arrangements, density, and overall structure. This gives pathologists a clear understanding of the tissue to differentiate tumor vs non-tumor regions. Once a tumor is detected, the prognosis is estimated based on the biological characteristics of the tumor. Immunohistochemistry (IHC) ~\cite{kim2016immunohistochemistry} staining is used to detect the behaviors that help labs to further divide the tumors into specific subcategories. IHC is a widely used technique in labs that uses antibodies to check for specific antigens. The most commonly used IHC Markers for breast cancer are ER, PR, HER2, and Ki67 in addition to H\&E (Tumor Detection).

\subsection*{Problems with conventional histopathology analysis}
Histopathology analysis involves examining tissue samples under a microscope to look for signs of cancer. The tissue samples are processed and placed on glass slides. Histopathologists look for changes in cell shapes and tissue distributions to determine if tissue regions are cancerous. Counting identified cells and then further classifying them based on their color is a very intensive task and brings subjectivity. Further adding to the complexity, Analysis with multiple markers is needed for efficient prognosis, predicting response to therapy, and evaluating residual tumor post-treatment. This multi-step process is laborious and very time-consuming. According to Diagnostic Concordance Among Pathologists Interpreting Breast Biopsy Specimens ~\cite{elmore2015diagnostic}, there is only 75\% concordance of diagnostic accuracy among professionals. Also, there is the problem of subjectivity and inter-observer variability which depends on the experience of the pathologist. Due to all the above problems, there was a growing need for automating and digitizing the pathological workflows.

With the advancement in technology, these glass slides were converted into whole slide images ~\cite{pantanowitz2010digital}, which can now be visualized on any digital screen. With the increasing popularity of digitization and reducing the cost of computing, the digitized slides have increased exponentially which has enabled pathologists and data scientists to utilize the power of advanced computing to better analyze the slides.

\subsection*{Related works}
Due to the high data requirement of deep learning algorithms~\cite{yasmin2013survey}, prior attempts for ROI Segmentation were made with traditional machine learning algorithms such as K-means clustering and SVM (Support Vector Machine) using hand-crafted feature extraction. However, these approaches were very time-consuming and required domain expertise. With the huge increase in digitization of slides, deep learning (DL)~\cite{shen2017deep} became one of the most popular tools for computer-aided diagnosis (CAD). DL architectures especially Convolutional Neural Networks (CNNs)~\cite{oshea2015introduction} have shown promising results in a lot of biomedical imaging tasks such as abnormality \& region detection, tissue \& organ segmentation, cell \& nuclei classification~\cite{xing2016robust}, image registration~\cite{cheng2016deep}, etc.

In almost all cases of CAD in cancer, one of the necessary preprocessing steps is tumor segmentation also known as Region of Interest (ROI) detection where each pixel is classified to separate the tumor from the rest of the normal tissue regions. This step reduces the processing time and computes for post-image analysis by reducing the target region and only focusing on the tumor region. Deep learning networks such as FCN~\cite{long2015fully}, SegNet~\cite{lbadrinarayanan2017segnet} \& UNet ~\cite{ronneberger2015u} are some of the well-performing semantic segmentation models. There are many publications summarized in~\cite{benny2021semantic} that have tried to automatically detect tumors in Breast cancer biopsy slides. Overall, UNet has proved to be the most popular network for biological segmentation. The limitation of these models is that they are stain-specific and mostly experimented with H\&E stain only. Also, these models are single-resolution models. So, they were not able to use the multi-resolution pyramidal structure of WSI slides. Due to this, the FOV of these models is fixed and they are not able to extract the local or global feature of the tissues based on their magnification. Another problem with these models is they only do tumor prediction/binary segmentation. But there are tissues such as acini, ducts, DCIS, etc. that can take up stains and look like tumor cells. Due to their similarity with Tumor regions, they are mostly detected as ROIs.

To resolve the single-resolution issue, a Multi-Resolution Network (MRN) was also published in~\cite{gu2018multi}. This paper used UNet architecture with multiple encoders for down-sampling and single decoders for up-sampling. All the decoder feature maps from low resolutions are concatenated with the corresponding decoder in high resolution. Due to this, the model was able to get both low and high-resolution FOV. The limitation of this model is that it center-crops the feature maps in lower magnification and then upscales the cropped feature maps during concatenation. This leads to limited usage of the feature maps from lower resolution.

Multiclass ROI Segmentation for H\&E stain was also attempted in ~\cite{ho2021deep}, where multiple models were trained with single \& multiple magnifications $(20\times, 10\times, 5\times)$ deep network. This model is a multi-encoder, multi-decoder, and multi concatenation architecture. This multi-magnification-based architecture achieves better performance than the single magnification model and even UNet and SegNet. But the problem with this model is they are not performing well-differentiated carcinomas and also, they are very sensitive to scanning quality. Similarly, HookNet ~\cite{rijthoven2021hooknet} is another multi-resolution multi-class deep learning model that was published for the lung cancer H\&E dataset. These models efficiently combine the feature maps across UNet branches giving special attention to the resolutions across encoders and decoders.

HER2 stain looks very different from other stains because this biomarker stains the membrane of the cells and not the nucleus. Publications such as ~\cite{dicataldo2010automated} used unsupervised color clustering and morphological features to automatically detect the cancerous regions in HER2 Stain. Trapezoidal LSTM was used for superior performance in HER2Net ~\cite{saha2018her2net} to fully automate the membrane and nucleus segmentation. In ~\cite{qaiser2019learning}, a Deep Reinforcement Learning model was used which sequentially identifies relevant ROI by following a parameterized policy. This model outperformed most of the state-of-the-art models of that time. The limitation of these models is that they are domain-specific and cannot be used for generic breast tissue segmentation.

\subsection*{Challenges associated with existing segmentation models for breast carcinoma}
There are many challenges in solving Medical Image Segmentation efficiently. These can be broadly classified into 2 categories:

\subsubsection*{Biological Issues}
Due to the lack of standardization, the process of tissue preparation involves multiple degrees of variability, e.g., in cutting thicknesses, staining protocols, dye compositions, etc. This leads to Inter and intra-laboratory stain variation. Moreover, there are regions on the slide that are not good for analysis like folds, out-of-focus areas, and bad regions. The current models are less confident in these regions, and it may consider those regions as ROI/tumor regions. After taking up stains, some tissues like acinis, Ducts, DCIS, Lymphocytes, etc. look very similar to tumor cells. So, they get detected as tumors by existing models. This misclassification generally leads to wrong Allred Scoring and hence improper prognosis.

\subsubsection*{Technical Issues}
On average, the slide dimensions at maximum resolution $(40\times)$ are $50000 \times 50000$ pixels. So complete slide processing in one go is not possible leading to the problem of the huge volume of data. Patch-wise or lower-resolution processing is used for training. But patch-wise processing leads to problems such as stitching issues at borders of patches, small FOV, and cropped tissue structures. For supervised training of deep learning models, vast labeled data is needed. Because of the enormous size of the slides, it becomes tedious and takes time to annotate the slides to generate ground truth for the model leading to insufficient labeled data issues. Different Levels of magnification give distinct levels of information in Whole Slide Images (WSI). So, it becomes critical to utilize multiple resolutions in getting a complete set of features for accurate segmentation. Single-resolution ROI segmentation models either fail to take in local or global features. Available Multi-Resolution ROI segmentation models are either very heavy on parameters because of the processing of 2 or more resolution levels or the context transfer between the resolutions is not efficient. Due to this, the models are not fully accurate on very large or very small tissues. Due to scanner settings \& optics across different scanners, color differences are noticeable in the digital display. These color variations are significant enough to change the output distribution of the model leading to stain variation issues. Most of the state-of-the-art models are mostly trained on one specific stain. This makes the training \& maintenance of the model difficult. Even though the morphological features of tumors remain the same across stains, there is no generic model that works across stains.

\subsection*{Methodology}
In this paper, we present a fully automatic multi-class tissue segmentation algorithm that classifies tumors as well as other tissue regions such as acini, ducts, DCIS, etc. into fine-grained segmentation maps of histopathology images. The proposed model works with multiple stains of breast IHC including but not limited to nuclei (ER, PR, Ki67) and membrane (HER2) stains. A huge volume of 10 million patches extracted from 513 WSI from multiple data sources and multiple scanners makes our model stain and scanner agnostic. Our model is immune to variations in slide preparation protocols and generic to the scanner properties. Our Multi-Encoder UNet Architecture with EfficientNetB4 makes use of the multi-resolution Pyramidal WSI extracting both local and global features with higher FOV. Due to the use of EfficientNet, our model is very lightweight with only 25 million Trainable parameters. Our unique merge logic between high-resolution and low-resolution branches efficiently combines pixel-aligned feature maps giving proper flow to context.

This paper proposes three novel approaches as follows:
\begin{enumerate}
    \item Single model detecting multiple tissue regions such as Tumor, acinis, ducts, DCIS, bad regions, etc. for multiple IHC markers including but not limited to nuclei (ER, PR, Ki67), cytoplasm (HE) and Membrane (HER2) staining.
    \item Novel multi-level multi-class UNet Architecture with unique Pixel-aligned Merge Logic between branches to support context flow without loss of any information.
    \item Unique multi-step training \& context-aware augmentation approach to handle partially annotated data and high-class imbalance.
\end{enumerate}

\begin{table}[!t]
    \centering
    \resizebox{\linewidth}{!}{%
    \begin{tabular}{c|c|c|c|c}
     \hline
    & Categorical Accuracy & Mean IoU & Weighted CE loss & Dice loss\\ \hline
       Base Model &  \begin{minipage}{.24\textwidth}
            \includegraphics[width=0.95\linewidth]{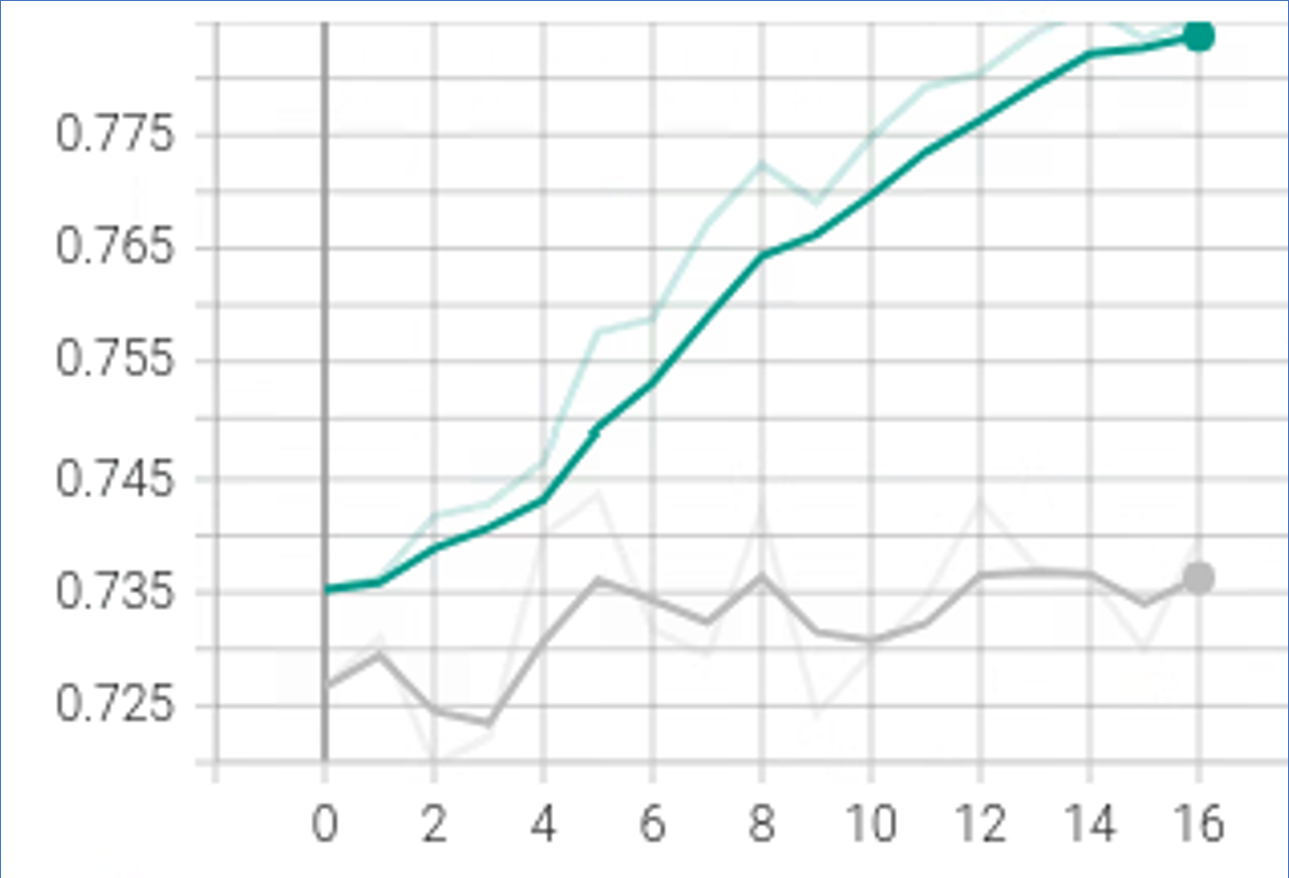}
            \end{minipage}  
          &
          \begin{minipage}{.24\textwidth}
            \includegraphics[width=0.95\linewidth]{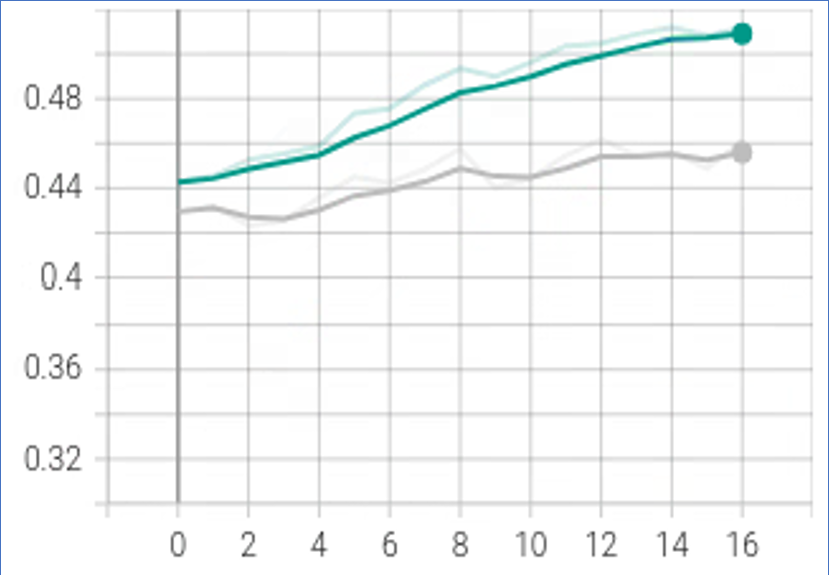}
            \end{minipage}  
           &
           \begin{minipage}{.24\textwidth}
            \includegraphics[width=0.95\linewidth]{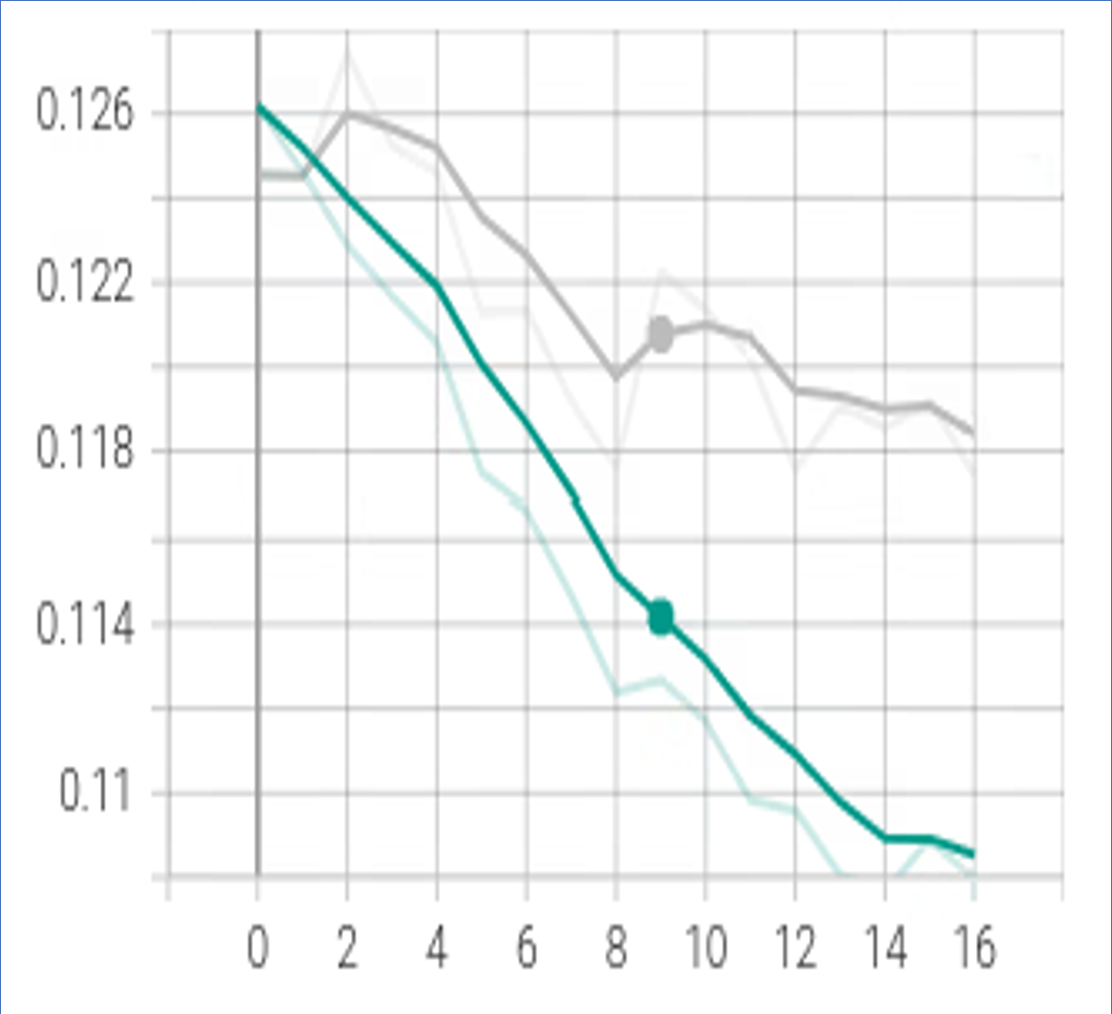}
            \end{minipage}
            &
            \begin{minipage}{.24\textwidth}
            \includegraphics[width=0.95\linewidth]{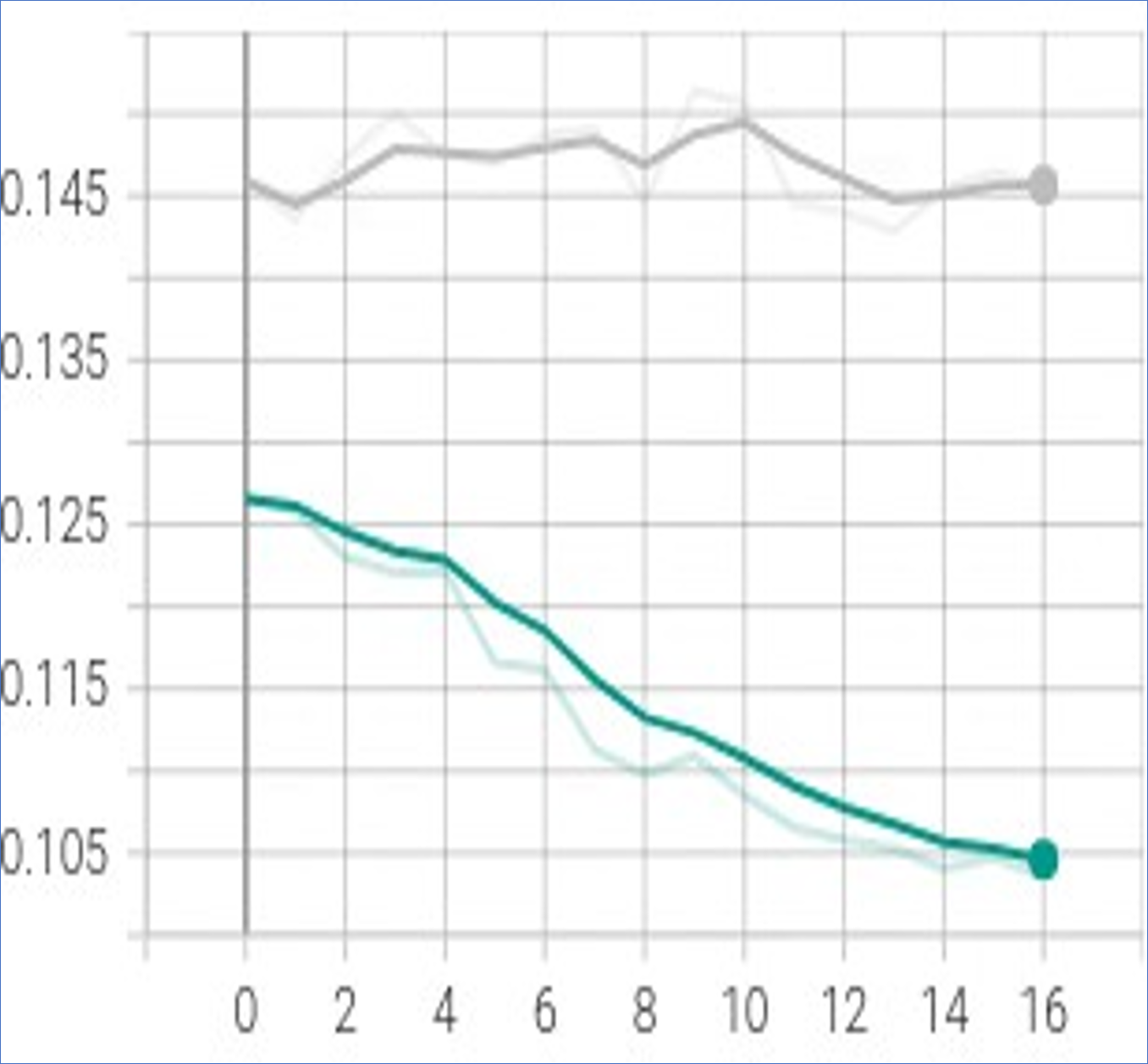}
            \end{minipage}  
            \\ \hline
      Master Model   & \begin{minipage}{.24\textwidth}
            \includegraphics[width=0.95\linewidth]{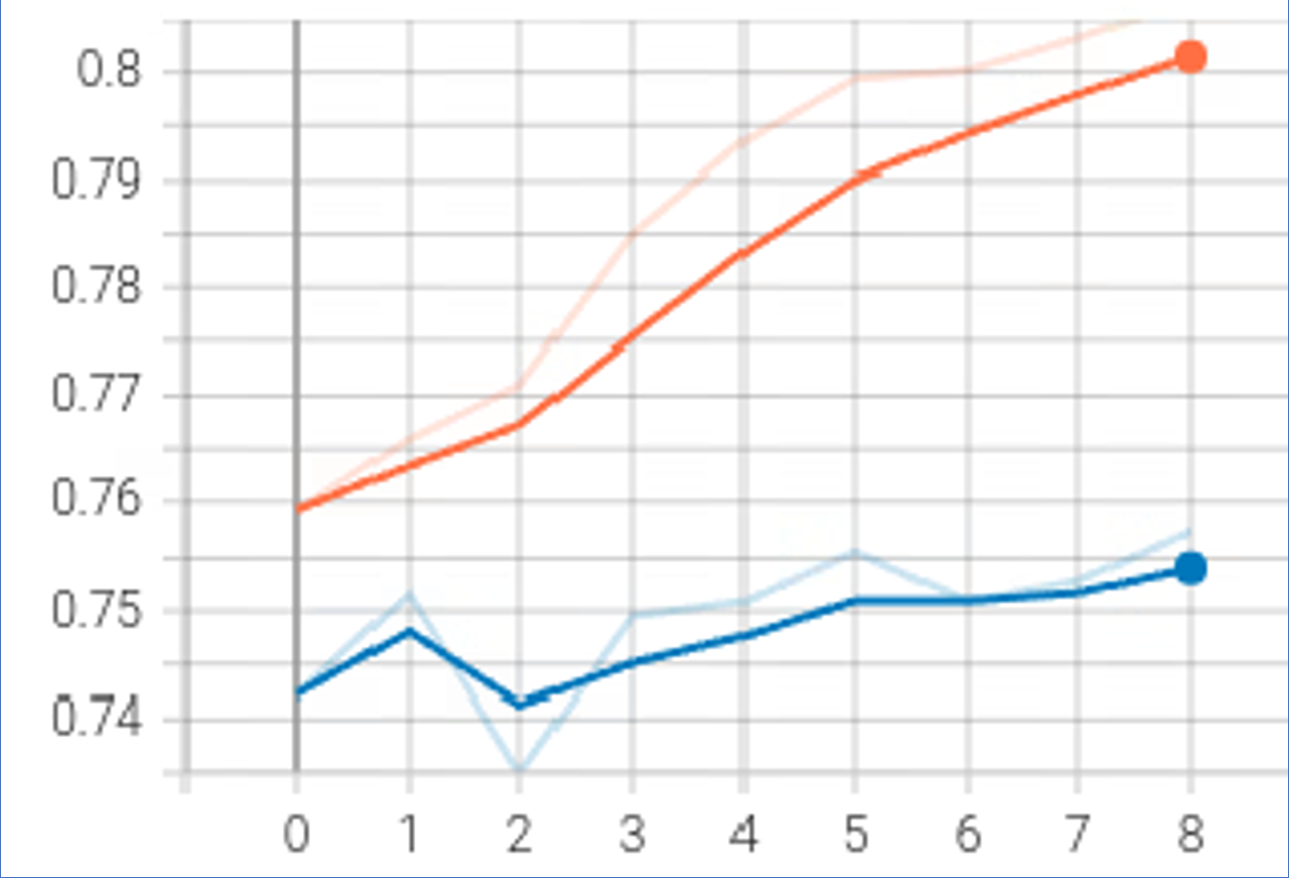}
            \end{minipage}
         & \begin{minipage}{.24\textwidth}
            \includegraphics[width=0.95\linewidth]{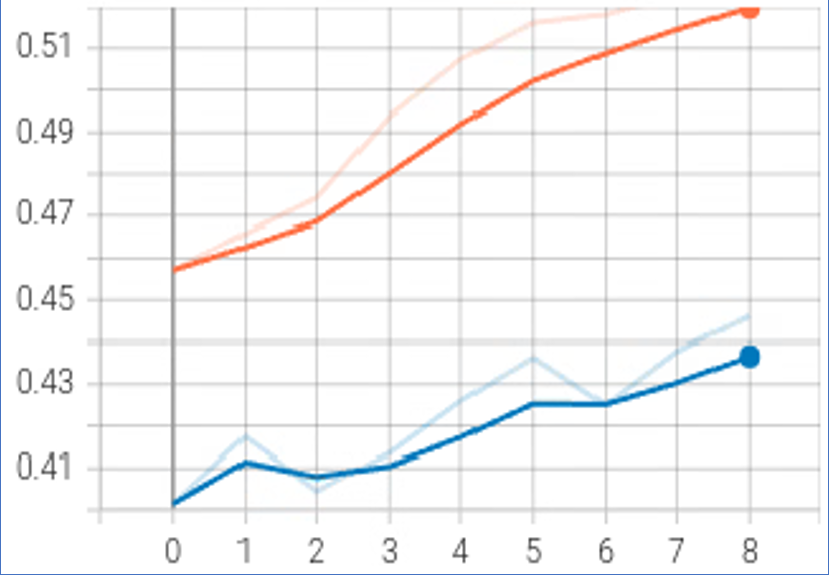}
            \end{minipage}
         & \begin{minipage}{.24\textwidth}
            \includegraphics[width=0.95\linewidth]{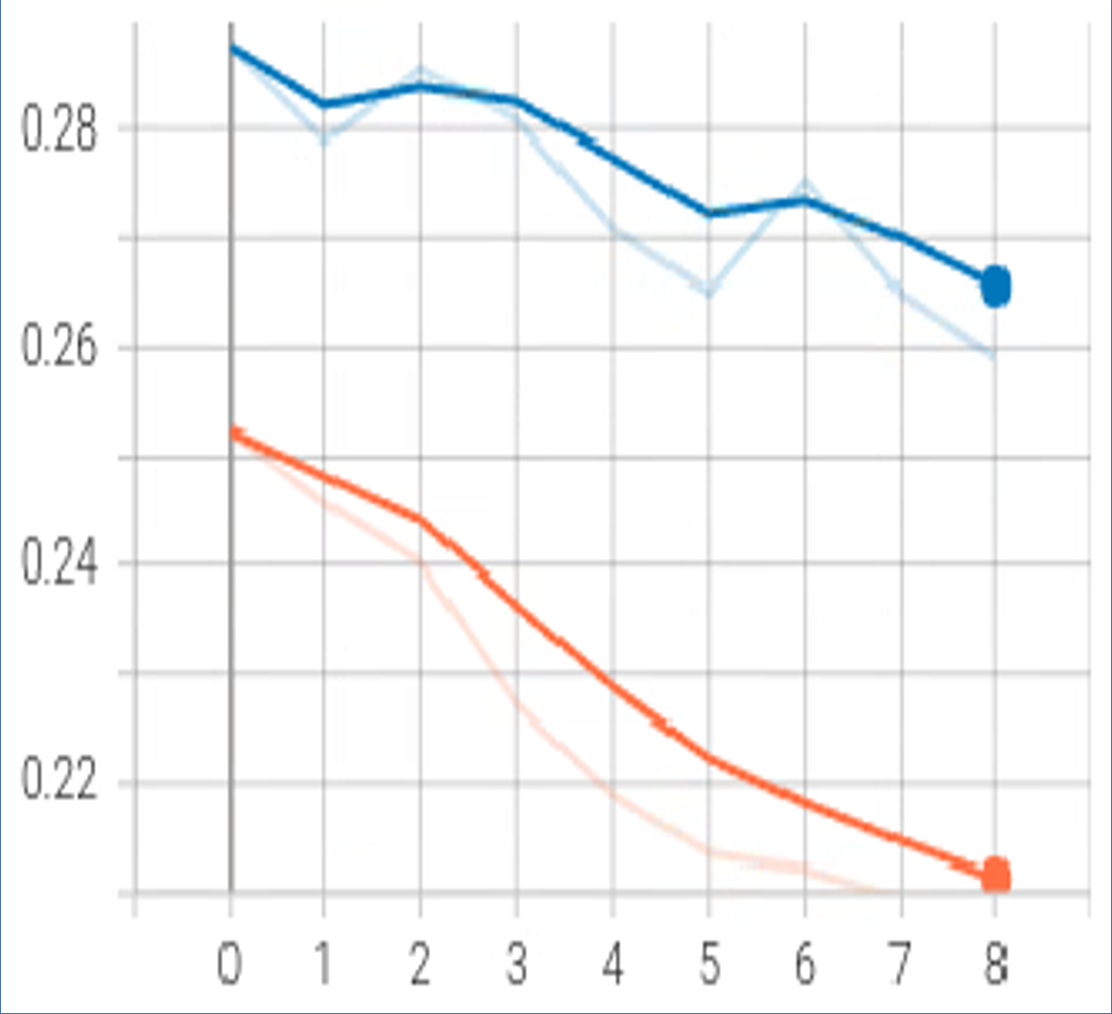}
            \end{minipage}
         & \begin{minipage}{.24\textwidth}
            \includegraphics[width=0.95\linewidth]{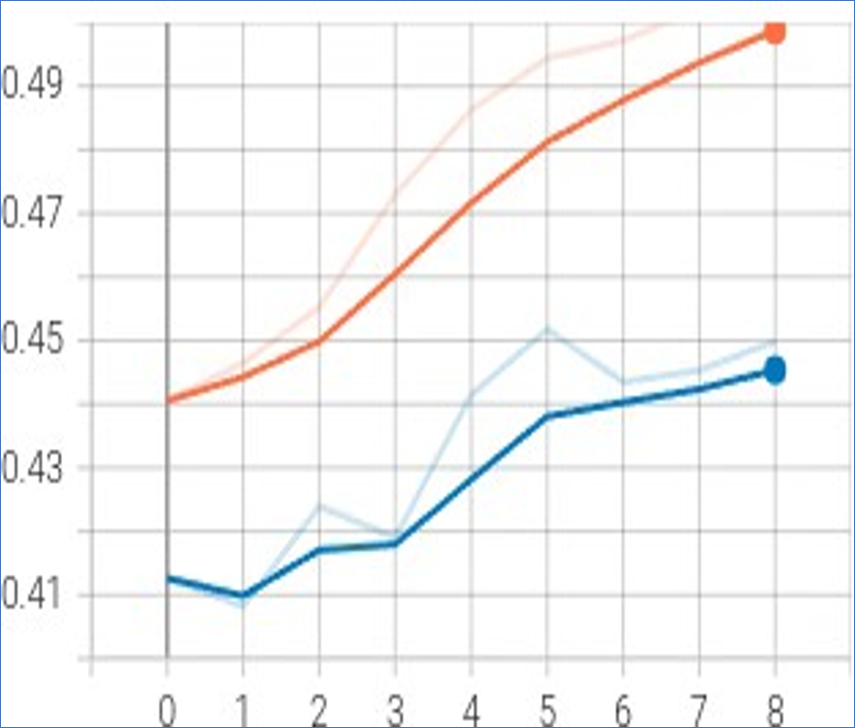}
            \end{minipage}  \\ \hline
    \end{tabular}%
    }
    \caption{Training \& validation accuracy, IOU, WCE \& dice loss provided by Tensorboard for the initial and the final step of the training. Green and Orange lines are the performance on the training set. Gray and blue lines are the performance on the validation set.}
    \label{tab:results}
\end{table}

\section*{Results}
In this study, we have used a total of 459 slides for training and 54 slides for validation. With augmentation, a total of 5 million patches were fed to the model for training. Table~\ref{tab:results} shows the improvement in model performance on training and validation set across epochs. The first row in Table~\ref{tab:results} gives the base model performance of 78\% Training accuracy and 73\% validation accuracy. The mean IOU (Intersection Over Union) across all channels is 0.50 for training and 0.44 for validation.

\begin{figure}[!t]
    \centering
    \subfloat[Tumor]{\label{fig:fig1_tumor}\includegraphics[width=0.32\linewidth]{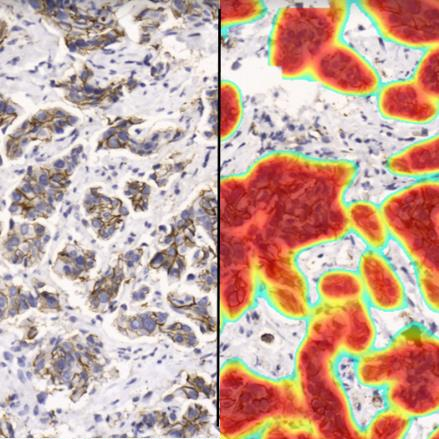}} \hfill
    \subfloat[DCIS]{\includegraphics[width=0.32\linewidth]{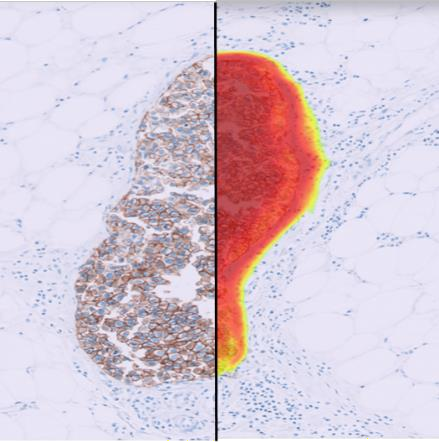}} \hfill
    \subfloat[Acini]{\includegraphics[width=0.32\linewidth]{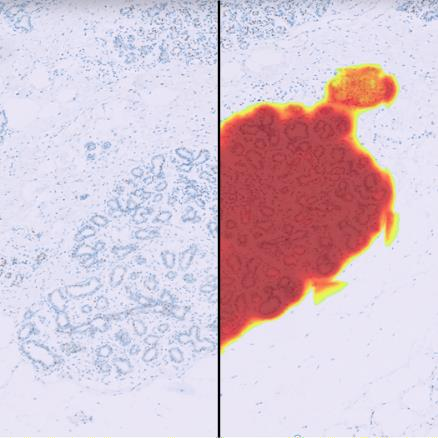}} \\
    \subfloat[Blood vessel]{\includegraphics[width=0.32\linewidth]{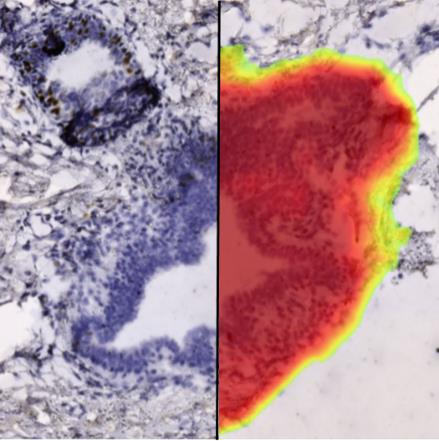}} \hfill
    \subfloat[Necrosis]{\includegraphics[width=0.32\linewidth]{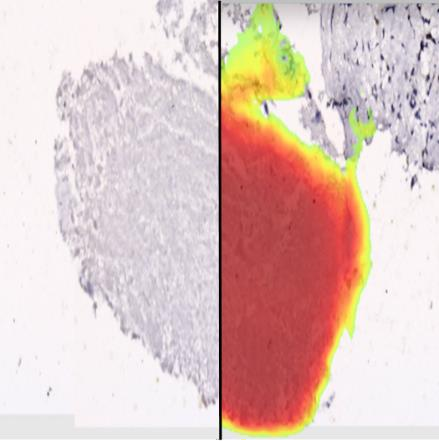}} \hfill
    \subfloat[Ducts]{\includegraphics[width=0.32\linewidth]{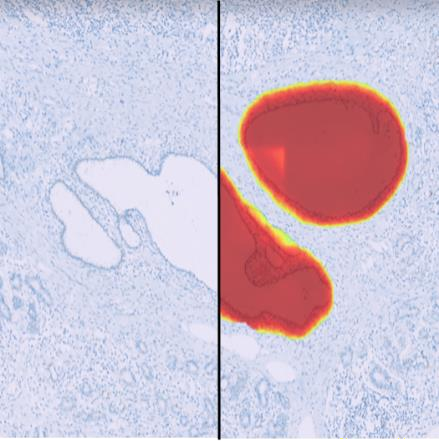}} 
    \caption{Heatmap prediction for different tissue regions on the test set of the dataset. The model's prediction confidence is represented by the intensity of the red region. The darker the intensity of red, more is the confidence of the model.}
    \label{fig:heatmap_prediction}
\end{figure}

There was a continuous decrease in weighted cross entropy loss and dice loss for both training and validation across epochs till epoch 12. After that losses flatlined showing that the model got saturated. The master model is iteratively fine-tuned for entire predicted datasets. The performance for all the iterations was almost similar to the final iteration with 80\% Training \& 75\% validation accuracy.

We have kept 55 slides, distributed equally across scanners, stains, and data sources that were not used in the training process(Detailed discussion in Dataset discussion). The Final Master model was validated in this dataset. Since the prediction is multiclass, so model was evaluated on 2 classes of tumors and others (including all classes other than tumor and normal). All the matrices are in terms of one class versus all other classes.

Our model prediction on different tissue regions can be seen in Figure~\ref{fig:heatmap_prediction}. Due to the use of context-aware merge logic, our model can perform well on small tumors also (refer to Figure~\ref{fig:fig1_tumor}). Our model is doing equally well on other tissue regions like DCIS, necrosis, ducts, acini’s, blood vessels, etc. with exceptionally smooth boundaries due to the utilization of weighted dice in the loss function. The model confidence is also extremely high as most of the valid tissue regions are in red color in Figure~\ref{fig:heatmap_prediction}.

Our proposed model has achieved a maximum IOU of 0.78 for tumors and 0.75 for other classes. The IOU between Ground Truth Annotation \& Model prediction is stable from threshold 0.2 to 0.8 before dropping steeply, showing the stability of the model and confidence in prediction across thresholds. The Precision \& Recall values can be referred to in Table~\ref{tab:performance}. They intersect at a threshold of approximately 0.4 for tumors and 0.45 for others. The optimum precision and recall value at threshold 0.4 is 0.83 for tumor \& 0.81 at threshold 0.45 for others. Similarly, the optimum sensitivity \& specificity value is 0.80 at 0.5 thresholds for tumors \& 0.90 at 0.3 thresholds for others. The values represent a good balance between false positives and false negatives making the model robust against both. Since all the intersections are in the range of 0.4 to 0.5, so we have taken 0.45 to be the final threshold for the classification of pixels. Detailed in Table~\ref{tab:performance} is another important metric AUC which is scale \& threshold invariant. The AUC is 0.85 for tumors \& 0.92 for other classes. This tells that model predictions are ranked well, and the quality of the prediction is also good.

Other than the test slides used for validation which were split from the training slides thus having a similar data distribution, we tested our model with a completely new stain (HE), Data source (DS7) \& Scanner (Motic) coming from real-time medical labs. The model has never seen these new distribution datasets during its training phase.

As seen in the highlighted region of Table~\ref{tab:iou_performance} the performance of the model on the new dataset was on par with the test dataset. The mean IOU of the tumor and other classes on 23 HE Stain was 0.73 \& 0.60 which is very similar to other stains as can be seen in the first column of figure Table~\ref{tab:iou_performance}. For data source DS7, the IOU ranges from 0.6 to 0.75 which is similar to other data sources. Similarly, the IOU range for Motic Scanner is 0.5 to 0.7 which is similar to other scanners. The performance of the model didn’t change much on testing for varied input distribution. This shows the robustness of the model towards stain, scanner \& data sources.

\begin{table}[!t]
    \centering
    \resizebox{\linewidth}{!}{%
    \begin{tabular}{c|c|c|c|c}
     \hline
    & IOU & Precision vs Recall & Sensitivity vs Specificity & AUC \\ \hline
       Tumor vs Rest &  \begin{minipage}{.24\textwidth}
            \includegraphics[width=0.95\linewidth]{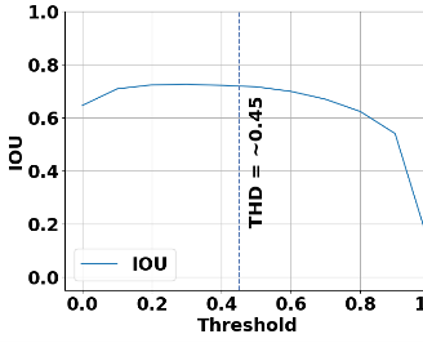}
            \end{minipage}  
          &
          \begin{minipage}{.24\textwidth}
            \includegraphics[width=0.95\linewidth]{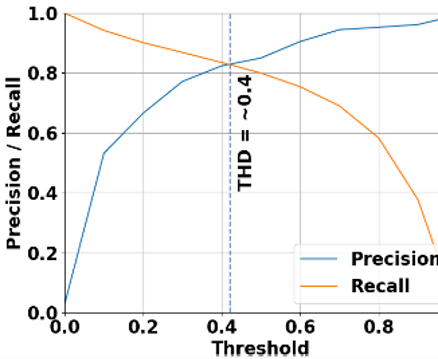}
            \end{minipage}  
           &
           \begin{minipage}{.24\textwidth}
            \includegraphics[width=0.95\linewidth]{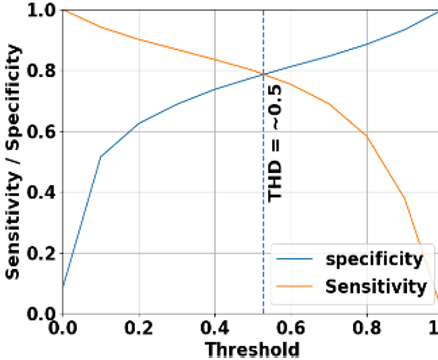}
            \end{minipage}
            &
            \begin{minipage}{.24\textwidth}
            \includegraphics[width=0.95\linewidth]{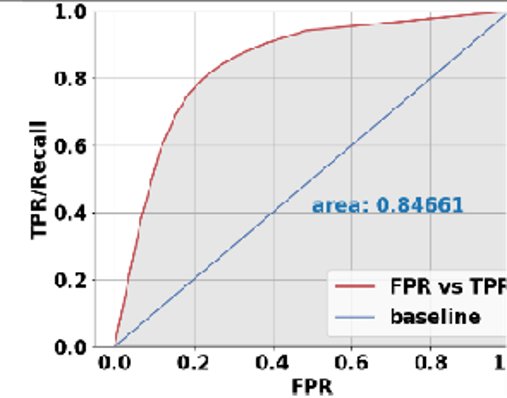}
            \end{minipage}  
            \\ \hline
      Other categories vs Rest   & \begin{minipage}{.24\textwidth}
            \includegraphics[width=0.95\linewidth]{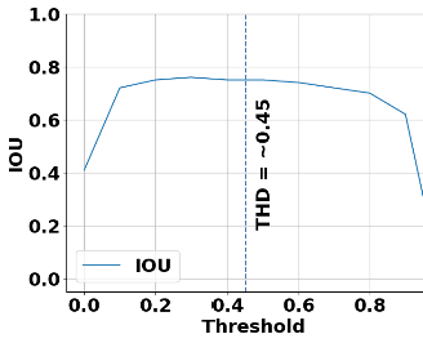}
            \end{minipage}
         & \begin{minipage}{.24\textwidth}
            \includegraphics[width=0.95\linewidth]{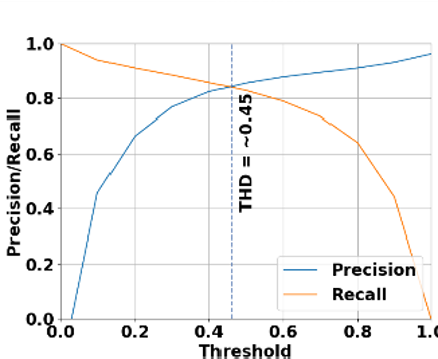}
            \end{minipage}
         & \begin{minipage}{.24\textwidth}
            \includegraphics[width=0.95\linewidth]{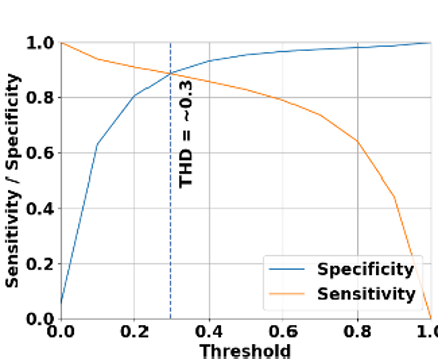}
            \end{minipage}
         & \begin{minipage}{.24\textwidth}
            \includegraphics[width=0.95\linewidth]{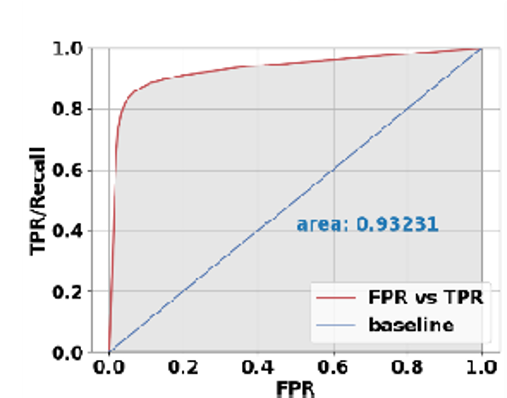}
            \end{minipage}  \\ \hline
    \end{tabular}%
    }
      \caption{Performance of final model. The first row has tumor vs everything else metrics. The second row has others vs everything else metrics. The values contained in columns left to right are IOU, precision, recall, sensitivity, specificity, and AUC Metrics, respectively}
    \label{tab:performance}
\end{table}

\begin{table}[!t]
    \centering
    \resizebox{\linewidth}{!}{%
    \begin{tabular}{c|c|c|c}
     \hline
    & Stain-wise IOU & Data source-wise IOU & Scanner-wise IOU  \\ \hline
       Tumor &  \begin{minipage}{.24\textwidth}
            \includegraphics[width=0.95\linewidth]{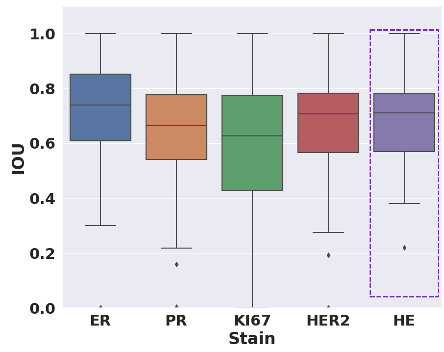}
            \end{minipage}  
          &
          \begin{minipage}{.24\textwidth}
            \includegraphics[width=0.95\linewidth]{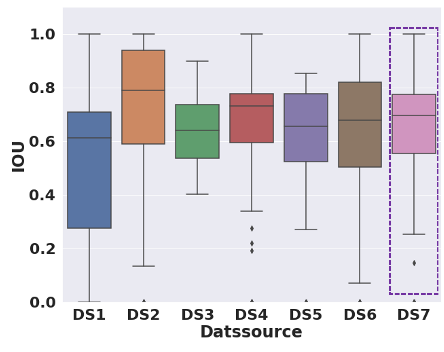}
            \end{minipage}  
            &
            \begin{minipage}{.24\textwidth}
            \includegraphics[width=0.95\linewidth]{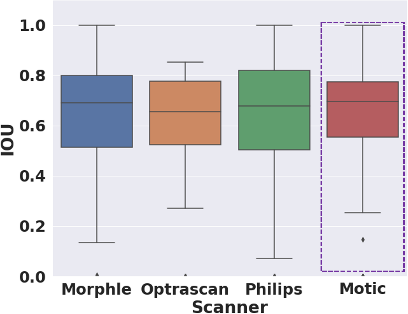}
            \end{minipage}  
            \\ \hline
      Others  & \begin{minipage}{.24\textwidth}
            \includegraphics[width=0.95\linewidth]{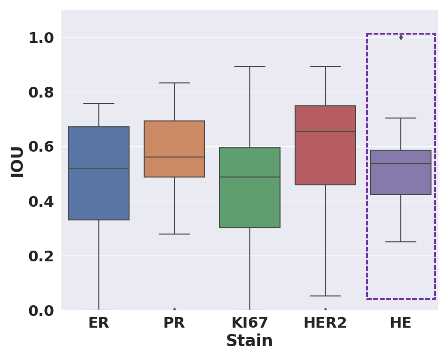}
            \end{minipage}
         & \begin{minipage}{.24\textwidth}
            \includegraphics[width=0.95\linewidth]{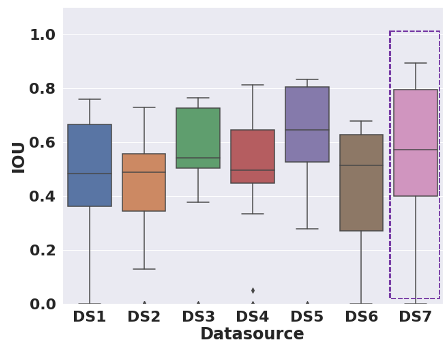}
            \end{minipage}
         & \begin{minipage}{.24\textwidth}
            \includegraphics[width=0.95\linewidth]{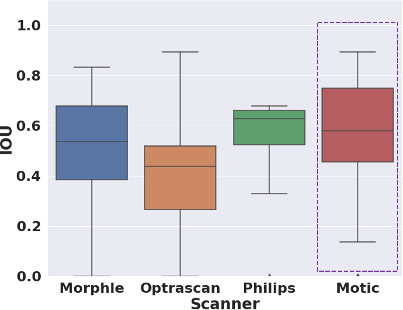}
            \end{minipage}  \\ \hline
    \end{tabular}%
    }
     \caption{Stain/Data source/Scanner wise IOU performance of the proposed model for tumor and other class. Highlighted boxes on all the plots are the model performance on the new input distribution dataset.}
    \label{tab:iou_performance}
\end{table}

\section*{Discussion}
In this study, we employed our novel multi-level deep learning architecture to accurately segment the tumor and other tissue regions such as acinis, duct, DCIS, folds, bad regions, etc. of breast cancer IHC slides. The multi-phase training approach with multiple augmentation practices enabled our model to learn varied contexts for the same and neighboring regions from partially annotated datasets. Our unique pixel-aligned merge architecture enabled the model for proper context transfer across UNet branches making the classification of difficult tissues such as ducts, acinis, DCIS, etc. possible. The heatmap in Figure~\ref{fig:heatmap_prediction} shows the model prediction of various tissue types with high confidence and smooth boundaries.

Our approach successfully addressed the current challenges of generic models for multiple stains across data sources. The high IOU for tumor and other classes in Table~\ref{tab:performance} clearly shows the model’s high performance. Our use of unique augmentation techniques such as heavy \& targeted augmentation has made our model scanner \& stain agnostic. Our use of GAN Style Transfer has made the model less variant towards texture and optical changes from multiple data sources. The concordance across the mean and range of the IOUs in the plots in Table~\ref{tab:iou_performance}, shows the model stability across stain, scanners \& data sources. The full potential of the model can be seen with validation on the new dataset coming from the Motic scanner, DS7 data source \& HE stain. Regardless of never getting trained on this input distribution, the model performed on par with the test dataset performance.

The proposed Model is supposed to be generic for nuclei and membrane stains across scanners and data sources. Currently, the model is validated on ER, PR, HER2, KI67 \& HE stains. In future works, we can validate this model for other stains such as P63, Cytokeratin 8, E-cadherin, etc. Currently, the model is validated for Breast IHC Slides, we can also validate this model for other parts of the body such as prostate, Colon IHC Slides, etc. Theoretically, the model should work on these cases too because of their tissue similarities and morphological learnings of the model. But in case of ambiguities, there is scope to improve the merging logic for a better unified mask. The current merging of UNet branches at the end is cropped and concatenated.



 



\section*{Methods}
\subsection*{Dataset}
A total of 513 Whole slide images (WSI) distributed across ER, PR, HER2 \& Ki67 stains are used for training \& validation. These datasets are from multiple diagnostic labs across different regions with different staining protocols. Each lab has used different scanners such as Philips, Morphle, Optrascan, etc. to digitize the slides. 55 Slides were set aside for testing. These slides were equally distributed across labs and stains. Please refer to Table~\ref{tab:total_data_distribution} for the exact distribution of data across stains/labs/Scanners.

\begin{table}[!t]
\centering
\begin{tabular}{lllllllll}
\hline
                            & Scanner & \multicolumn{2}{c}{Morphle} & \multicolumn{2}{c}{Optrascan} & \multicolumn{2}{c}{Philips} &       \\ \hline
Slides                      & Stain   & DS1          & DS2          & DS3           & DS4           & DS5          & DS6          & TOTAL \\ \hline
\multirow{5}{*}{Training}   & ER      & 21           & 3            & 15            & 16            & 12           & 18           & 85    \\ \cline{2-9} 
                            & PR      & 34           & 22           & 17            & 15            & 9            & 18           & 115   \\ \cline{2-9} 
                            & HER2    & 55           & 17           & 19            & 17            & 15           & 19           & 142   \\ \cline{2-9} 
                            & Ki67    & 41           & 15           & 20            & 13            & 7            & 21           & 117   \\ \cline{2-9} 
                            & TOTAL   & 151          & 57           & 71            & 61            & 43           & 76           & 459   \\ \hline
\multirow{5}{*}{Validation} & ER      & 2            & 1            & 2             & 2             & 1            & 2            & 10    \\ \cline{2-9} 
                            & PR      & 5            & 4            & 3             & 1             & 1            & 1            & 15    \\ \cline{2-9} 
                            & HER2    & 6            & 2            & 1             & 2             & 2            & 3            & 16    \\ \cline{2-9} 
                            & Ki67    & 5            & 1            & 1             & 2             & 3            & 1            & 13    \\ \cline{2-9} 
                            & TOTAL   & 18           & 8            & 7             & 7             & 7            & 7            & 54    \\ \hline
\multirow{5}{*}{Test}       & ER      & 2            & 1            & 3             & 2             & 2            & 3            & 13    \\ \cline{2-9} 
                            & PR      & 2            & 2            & 3             & 3             & 2            & 3            & 15    \\ \cline{2-9} 
                            & HER2    & 2            & 1            & 3             & 1             & 2            & 3            & 12    \\ \cline{2-9} 
                            & Ki67    & 2            & 2            & 2             & 4             & 2            & 3            & 15    \\ \cline{2-9} 
                            & Total   & 8            & 6            & 11            & 10            & 8            & 12           & 55    \\ \hline
\end{tabular}
\caption{Stain-wise, scanner-wise wise \& data source-wise training, validation \& test set distribution.}
\label{tab:total_data_distribution}
\end{table}

Ground truth annotations for these slides were done by a panel of 2 senior and 2 junior pathologists and quality controlled by domain experts. Since complete slide annotation is very difficult and time-consuming, a selective region-wise annotation technique was utilized. Pathologists select 2-3 mutually exclusive non-similar small regions and annotate completely with multiple labels such as Tumor, Normal, Stroma, acini, duct, Blood vessels, DCIS, etc. Annotations for bad regions such as Folds, Artifacts, bubbles, out-of-focus areas, etc. were also done so that we could detect these areas properly. We would extract patches from only these annotated regions. This approach helped us in getting sufficient distribution of data with limited time and resources.

\subsection*{Class Imbalance}
\begin{figure}[!b]
    \centering
    \subfloat[Training set]{\includegraphics[width=0.49\linewidth]{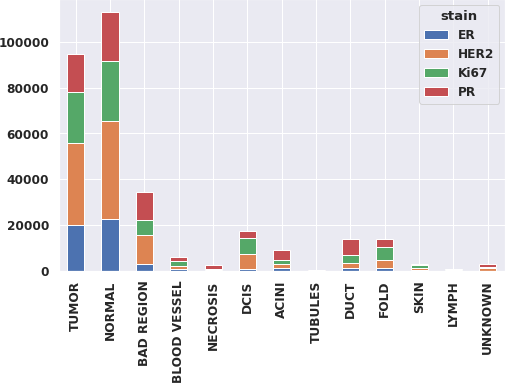}} \hfill
    \subfloat[Validation set]{\includegraphics[width=0.49\linewidth]{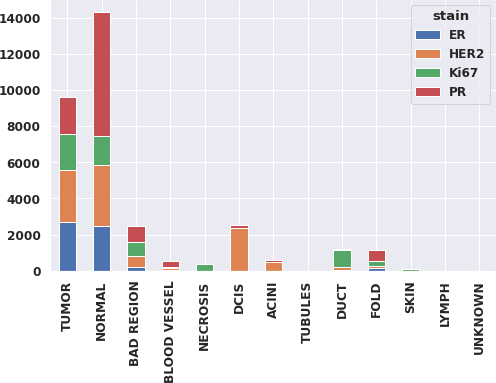}}
    \caption{The distribution of different tissue categories across (a) training set, and (b) validation set.}
    \label{fig:class_imbalance}
\end{figure}

High-class imbalance is one of the major problems in segmentation tasks. In most of the WSI, Tumor regions are far less as compared to normal tissue regions and background. Using the Otsu filter, we ignore the non-tissue area and background. As a majority of the annotated tissue regions are either Tumor or Normal/Stroma tissues (refer to Figure~\ref{fig:class_imbalance}), so we have clubbed Bad Regions, Blood Vessels, DCIS, acini, ducts, Tubules, Folds, Skin, Lymph \& Unknowns in a separate class known as Others. This reduces the imbalance of specific tissue types concerning tumors. Table~\ref{tab:data_distribution} describes the patch label distribution of each class after clubbing. Approximately 300,000 raw training patches \& 33,000 raw validation patches were extracted from slides.

\begin{table}[!t]
\centering
\begin{tabular}{llccc}
\hline
Class & Label                                    & \multicolumn{1}{l}{Raw Training Patches} & \multicolumn{1}{l}{Augmented Patches} & \multicolumn{1}{l}{Validation Patches} \\ \hline
0     & Background/Normal                        & 113,110                                  & 1,809,760                             & 14,325                                 \\ \hline
1     & Tumor                                    & 94,708                                   & 1,515,328                             & 9,597                                  \\ \hline
2     & Normal, Stroma, Acini, Duct, DCIS   etc. & 103,350                                  & 1,653,600                             & 8,692                                  \\ \hline
Total &                                          & 311,168                                  & 4,978,688                             & 32,884                                 \\ \hline
\end{tabular}
\caption{Class-wise distribution Training and validation Patches}
\label{tab:data_distribution}
\end{table}

Another kind of imbalance occurs if random sampling of patches is used for training the model. Most of the batches get an imbalanced distribution of patches. To resolve this issue, our data generator pipeline ensures that all batches have the class distribution of 2:1:1 (tumor: normal: others).

\begin{figure}[!t]
    \centering
    \includegraphics[width = 0.95\linewidth]{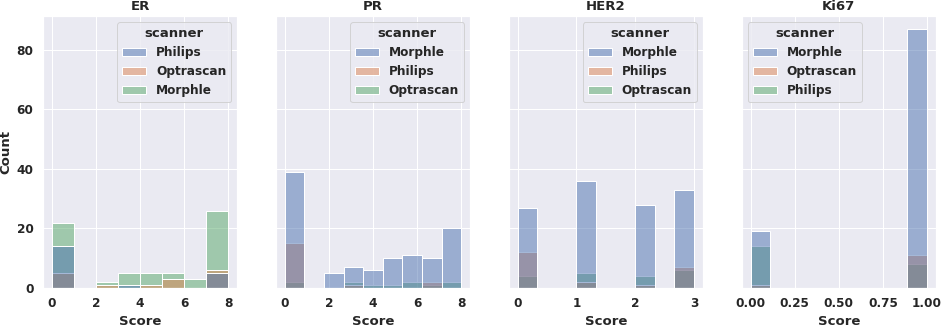}
    \caption{Scores distribution across stains ER, PR, HER2 \& Ki67 vertically stacked with Philips, Morphle \& Optrascan scanners.}
    \label{fig:score_distribution}
\end{figure}

During exploratory data analysis (EDA), we found that despite uniformly distributed batches, there is a pixel-level imbalance problem among classes. This occurs because, at the batch level, the non-majority class pixel in each patch adds up to become the majority pixel. This degrades the performance of the gradient descent because at each step the model gets a different distribution of input samples. We have countered this problem by multiplying the dynamic weights $W^D_C$ of class c (calculated based on the count of the pixel label $Y$ for each patch of dimensions $w \cdot h$ present in a batch $b$ ) by the loss of each class. Equation~\ref{eq:dymanic_weight} gives the formula for dynamic weight calculation.

\begin{equation}
    W^D_C = 1 - (\sum_{j=0}^{b}~\sum_{i=0}^{w~\cdot~h} Y_{ij}^c\slash {\sum_{k=0}^{c}\sum_{j=0}^{b}~\sum_{i=0}^{w~\cdot~h}} Y^k_{ij})
    \label{eq:dymanic_weight}
\end{equation}

The dynamic weights are normalized so that the losses don’t get scaled much. Since the dynamic weights of classes are inverse of their presence, multiplying them with the class losses equalizes the impact of loss for the low occurring class.

\subsection*{Variance in Data}
As referred to in Table~\ref{tab:total_data_distribution}, The dataset used for our experiment is from 3 different scanners, 4 different stains, and 6 different data sources with different scanning protocols. This section summarizes the variance in the distribution of data across scores, stains, scanners \& data sources.

ER/PR generally follows Allred scoring standard which ranges from 0-8. HER follows CAP Guidelines which range from 0-3 and Ki67 ranges from 0-100\% but is mostly divided into 2 categories $<15$ and $\geq 15$. Figure~\ref{fig:score_distribution} gives the stain-wise score histogram of the training dataset. It shows that most of the sides in all the stains are either from high or low scores. These are quite easy slides to process because of no or extremely high staining. However, the difficult slides (moderate staining) are low in count.

\begin{figure}[!t]
    \centering
    \includegraphics[width=0.65\linewidth]{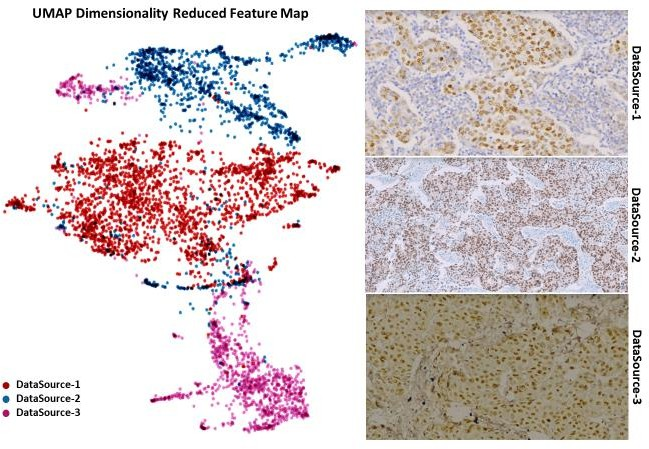}
    \caption{Data Variations across Multiple data sources. The right side of the image is sample patches and the left side of the image is 2D feature-reduced clusters from multiple data sources.}
    \label{fig:data_variation}
\end{figure}

\begin{figure}[!t]
    \centering
    \includegraphics[width=0.70\linewidth, height=0.4\linewidth]{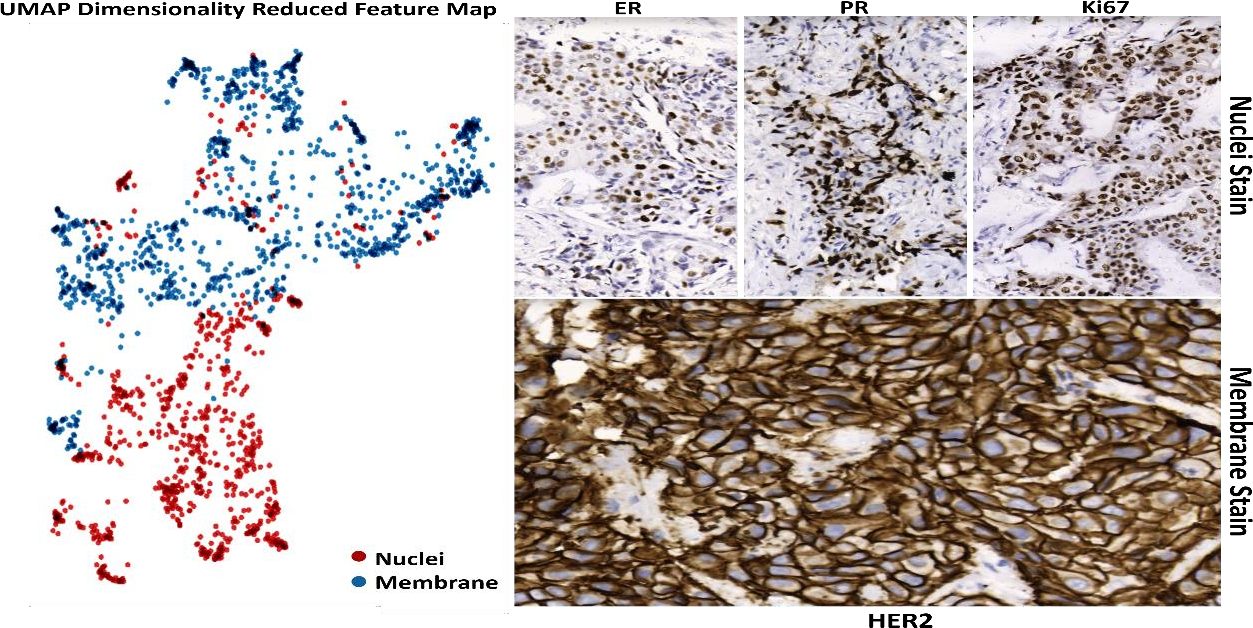}
    \caption{Variations across Nuclei (ER, PR, Ki67) \& Membrane (HER2) stain. The right side of the image is sample patches and the left side of the image is 2D Feature reduced clusters of nuclei \& membrane stains.}
    \label{fig:variation_across_nuclei}
\end{figure}

Figure~\ref{fig:data_variation} shows the variation in the distribution of data across data sources. Clear visual \& texture differences can be seen in membrane and nuclei staining. Figure~\ref{fig:data_variation} also shows the clustering of reduced feature maps in 2D. The feature maps were extracted by passing patches from the Reset50 network~\cite{he2016deep} with ImageNet weights. 2048 feature vector extracted from Resnet was reduced to a 2D feature map using Uniform Manifold Approximation and Projection for Dimension Reduction (UMAP). UMAP~\cite{mcinnes2018umap} uses graph layout algorithms to arrange data in low-dimensional space. It first constructs the high dimensional graph using a “fuzzy simplicial complex”. This is just a representation of a weighted graph, with edge weights representing the likelihood that two points are connected. Once the high-dimensional graph is constructed, UMAP optimizes the layout of a low-dimensional analog to be as similar as possible. Figure~\ref{fig:variation_across_nuclei} shows completely separable clusters created by UMAP (1 for nuclei and 1 for membrane). This shows the difference in data distribution for both kinds of stain.

Figure~\ref{fig:variation_across_nuclei} shows patches from multiple data sources with varying brightness, contrast, texture, and color. A clear difference can be seen in the brown nuclei staining and surrounding stroma. Patch from data source- 3 can be seen in mostly yellowish brown whereas sparse brown and solid brown in data source-2 \& data source-1 respectively. This kind of difference occurs due to variations in the staining protocols of different labs. Figure~\ref{fig:variation_across_nuclei} also shows dimensionally reduced feature maps of patches from different data sources projected in the 2D plane. Three separable local clusters can be seen for 3 data sources showing the clear difference in data distribution across data sources.

\begin{figure}[!t]
    \centering
    \includegraphics[width=0.65\linewidth, height=0.4\linewidth]{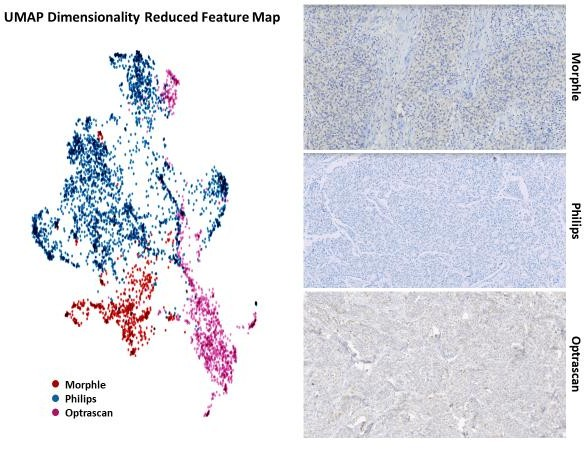}
    \caption{Data Variations across scanners. The right side of the image is sample patches and the left side of the image is 2D feature-reduced clusters across Philips, Morphle \& Optrascan scanners.}
    \label{fig:variation_across_scanner}
\end{figure}

Figure~\ref{fig:variation_across_scanner} shows patches from multiple scanners used in our experiment. Variations can be seen in the Intensity, color range, clarity, and quality of the patches across Morphle, Philips \& Optrascan. These variations occur due to different Optical quality, color calibration, resolution, and scanning speed used by different scanners. These variations result in variation in data distribution which can be seen in dimensionally reduced 2D projected feature maps of the patches across scanners. Three local clusters can be seen for three different scanners.

As we have seen in previous paragraphs, there are noticeable variations in the data distribution across scanners, stains, and data sources. So, the generic model needs to learn from all these variations to be a data source, scanner \& stain agnostic.

Most of the current state-of-the-art models are trained with only one stain or scanner. Hence, they fail to generalize on the other distributions. We have resolved these problems using various augmentations (discussed in the next section). This makes our proposed model generic across multiple stains, scanners, and data sources.

\subsection*{Augmentations}
Data Augmentation is the most common technique in machine learning and computer vision, especially for tasks like image classification, object detection, and image segmentation. The purpose of data augmentation is to increase the diversity of the training data without collecting additional samples manually, which can help improve the performance and robustness of machine learning models. It enhances performance by decreasing the variance in the model i.e. the model becomes much more insusceptible to real-world test data, improving the robustness and generalization of the model. During data preparation, it was ensured that the augmented data distribution was not skewed from the original.

Due to privacy concerns and regulations, medical Image datasets are much sparser in number in comparison to other text and image datasets. Due to various data augmentation, our training dataset became 16 folds, increasing the number of training patches approximately to 5 million (refer Table~\ref{tab:data_distribution}). We have not used augmentations only to increase the dataset but also to solve various other issues like color normalization, stain \& scanner invariance, and data source invariability. Explained below are the various augmentations used during training.

\begin{figure}
    \centering
    \includegraphics[width=\linewidth]{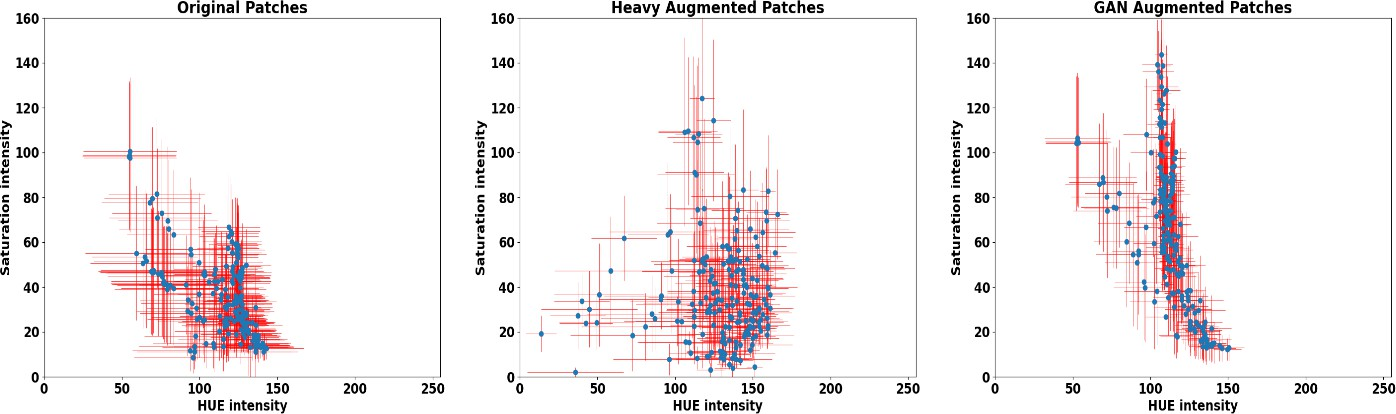}
    \caption{Effects of Augmentation on input patches. The first column is the hue-saturation range for original patches. The second column is the hue-saturation range for heavily augmented patches. The last column is the hue-saturation range for GAN-synthesized patches.}
    \label{fig:augmentation_effect}
\end{figure}

The effects of heavy augmentation and StyleGAN augmentation can be seen in Figure~\ref{fig:augmentation_effect}. The middle row shows spread in hue \& Saturation after applying heavy augmentation while the style shrinks the color range to the target range.

\textbf{StyleGAN Augmentations}~\cite{karras2019style} - StyleGAN is an extension of generative neural network architecture that uses two new sources of randomness to generate a synthetic image. We have trained Multiple StyleGAN models from scratch with input patches from one data source/scanner and target patches from other data sources/scanners. We have used the SSIM~\cite{hore2010image} metric as a loss while training to ensure that GAN-generated tissue looks structurally the same as input tissue. It is a perceptual metric to measure the similarity of two images by accounting for the luminance, contrast, and texture of the images. In Equation~\ref{eq:ssim}, $\mu_x$ and $\mu_y$ are pixel means computed with a Gaussian filter at a pixel with standard deviation $\sigma_{xy}$. $C_1$ and  $C_2$ are constants

\begin{align}
SSIM(p) = \frac{2\mu_x \mu_y + C_1}{\mu_x^2 + \mu_y^2 + C_1} \cdot \frac{2 \sigma_{xy + C_2}}{\sigma_x^2 + \sigma_y^2 + C_2}
    \label{eq:ssim}
\end{align} 
The model takes the input patch from the source data source and converts the style with the target data source. We infuse these pre-generated patches with a probability of 0.3 into training. This helps our proposed model to accommodate the variance in data distributions coming from different scanners (refer to Figure~\ref{fig:variation_across_scanner}) and data sources (refer to Figure~\ref{fig:variation_across_nuclei}).

\begin{figure}[!t]
    \centering
    \includegraphics[width=0.85\linewidth]{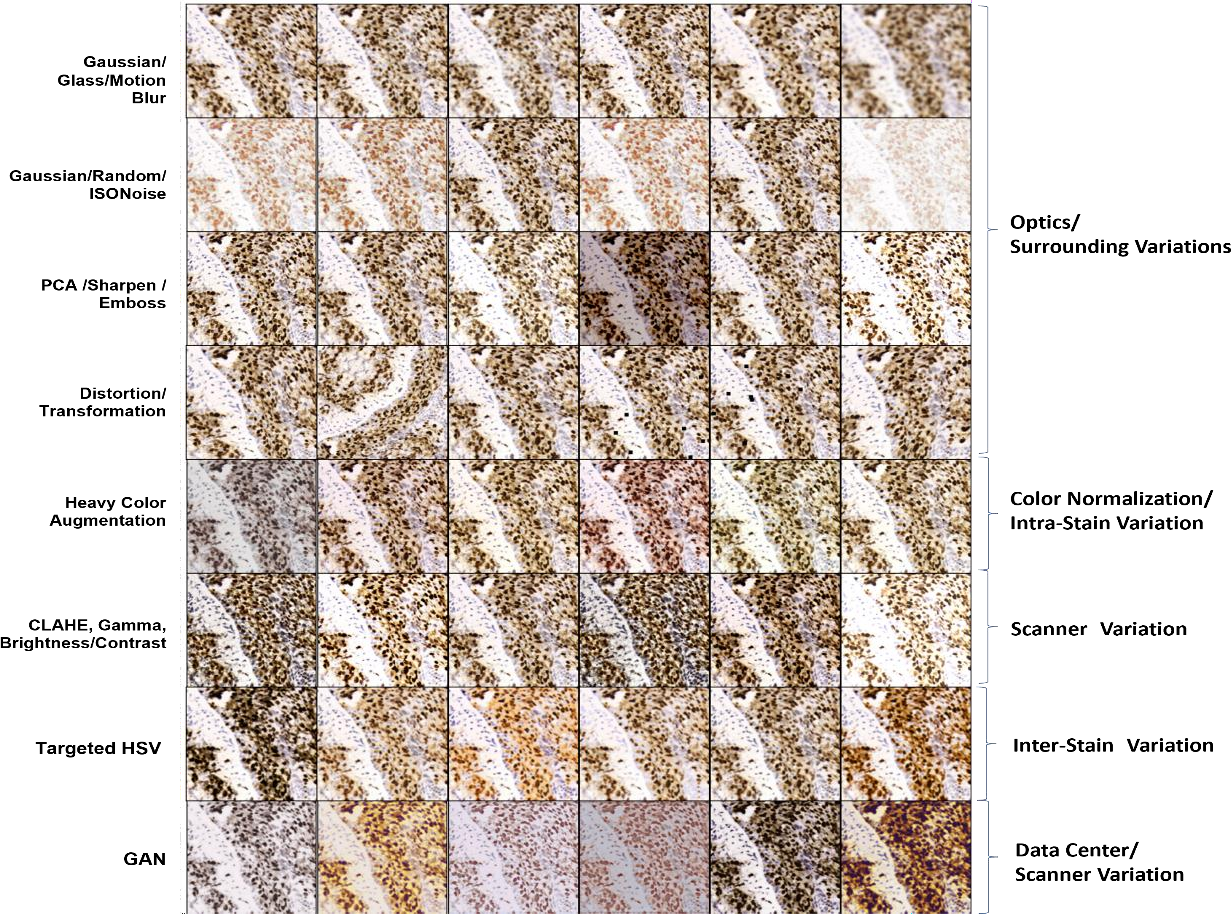}
    \caption{Augmentation techniques used to encounter stain/scanner/datasource variation. The left column shows the augmentation type and the right side shows the value addition it brings to the model. The center shows the multiple variations of generated patches by augmentation.}
    \label{fig:augmented_samples}
\end{figure}

\textbf{Heavy color augmentation} - Most of the state-of-the-art approaches use color normalization as a preprocessing step to encounter the problem of stain variance due to different staining protocols. This makes the model heavy due to the additional layers for normalization at the beginning of the model refer~\cite{tellez2019quantifying} for detailed implementation. So instead of normalizing the patches, we have used a heavy color augmentation (refer to Figure~\ref{fig:augmented_samples} row 5) technique where the patches are augmented with extreme colors. Due to large color variation, the model adapts to ignore the colors and focus on the morphology of the tissues. Hence removing the need for color Normalization. This technique makes the model robust with different stains such as ER, PR, HER2 \& Ki67.

\textbf{Targeted HSV Augmentation} - To make our model more robust towards the stain invariance, we needed more variance of color in only nuclei regions of the patches and not the background. All the current augmentation techniques are applied over the entire patch equally. This makes changes to the unstained background and stroma regions also. But due to staining variations only stained regions differ. Using the HUE channel of the HSV color space, we were able to separate the nuclear region from the background. Analyzing the nuclear region of the patches available, we found the valid saturation \& value range for blue and brown colors. (Refer to Table~\ref{tab:saturation}). Using this range, we were only augmenting the nuclear region for brown and blue variations mimicking the behavior of real-time stain variations. In the figure, row “Targeted HSV” shows the variations in stained regions.

\begin{table}[]
\centering
\begin{tabular}{lcc}
\hline
      & \multicolumn{1}{l}{Saturation} & \multicolumn{1}{l}{Value} \\ \hline
Brown & {[}-30,81{]}                   & {[}-80, 81{]}             \\ \hline
Blue  & {[}-40, 61{]}                     & {[}-60, 61{]}             \\ \hline
\end{tabular}
\caption{Saturation/Value range in HSV color space for brown and blue color}
\label{tab:saturation}
\end{table}

\subsection*{Transformation and Other Augmentations}
To build a robust model, we have used a variety of other Image transformations as shown in Figure~\ref{fig:augmented_samples}. We have used rotate, flip, crop, rescale, shift, and xy-jitter to increase the number of patches and provide multiple surrounding contexts of the same patch to the model. This helps the model to learn the boundaries of the patches better. We have used Gaussian/Glass/Motion Blur which tries to simulate optical variation occurring due to image capture equipment like cameras. Gaussian/Random Noise to simulate variations due to imaging modalities \& PCA/Emboss to highlight the edges in the scans. The above augmentations help the model account for variance due to environment and optics. We have used transformations such as flip, rotate, shear, crop, and rescale to simulate variations in positioning and orientation. Local elastic transformations \& Grid/Optical distortions were used to accommodate organ and tissue deformation like folds, cell clustering \& clotting. These augmentations increased the input patches more than 10-fold. We have used CLAHE, Gamma, Brightness \& Contrast to accommodate variation in the slides due to scanners such as Morphle, Optrascan, Philips, etc.

\subsection*{Model}
As discussed in the related works section, many architectures are trying to solve the segmentation tasks. We tried some of those architectures, but the performance was less than 0.5 Intersection Over Union (IOU). DeeplabV3 was unable to scale across multiple stains. So different models needed to be finetuned for different stains. Also, the predicted boundaries of this model were not smooth. The Vanilla UNet Model was performing better than deeplabV3 in terms of multi-stain prediction but was not able to do well for multi-class prediction because of the similarity between classes. The tumor class needs a local high-resolution Field of View (FOV) whereas the detection of other classes such as acini, ducts, DCIS, etc. needs global context/higher FOV. HookNet~\cite{rijthoven2021hooknet} solved this problem by taking multiple resolutions of patches with multiple UNet in parallel. However, the context transfer from low-resolution to high-resolution branch was not adequate. This resulted in high resolution doing better for tumor class whereas low resolution doing better for other classes.

\begin{figure}[!t]
    \centering
    \includegraphics[width=\linewidth]{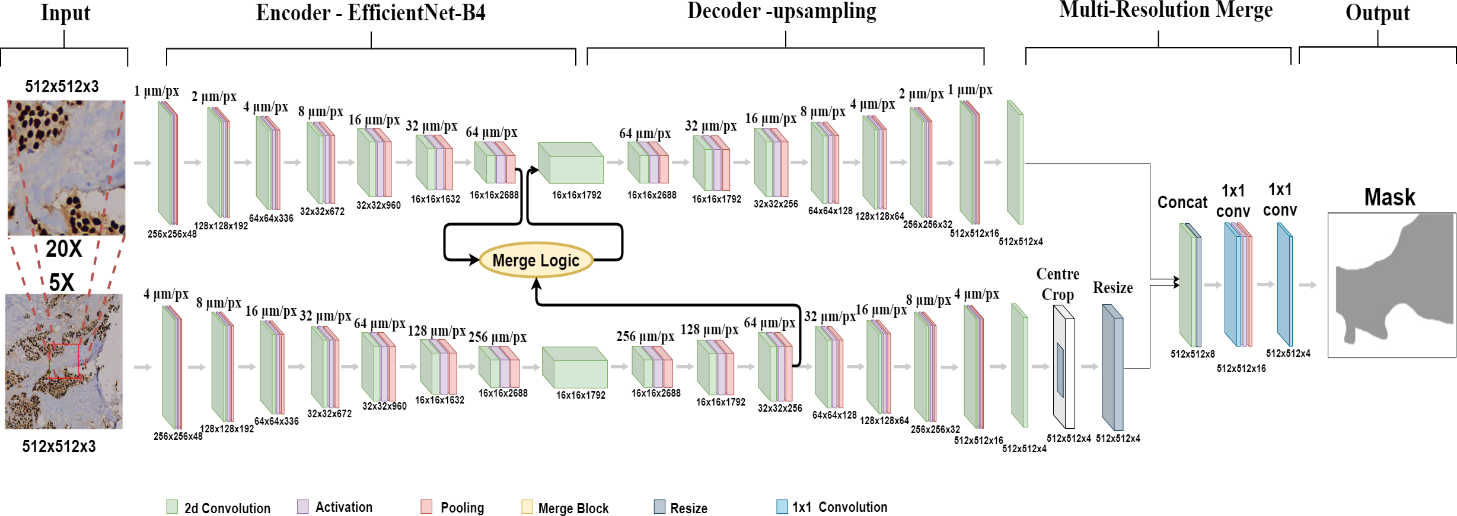}
    \caption{Our Novel Multi-class Multi-level UNet architecture with unique merge-logic and final concatenation layer to generate a single mask.}
    \label{fig:model}
\end{figure}

\begin{figure}[!t]
    \centering
    \includegraphics[width=0.65\linewidth]{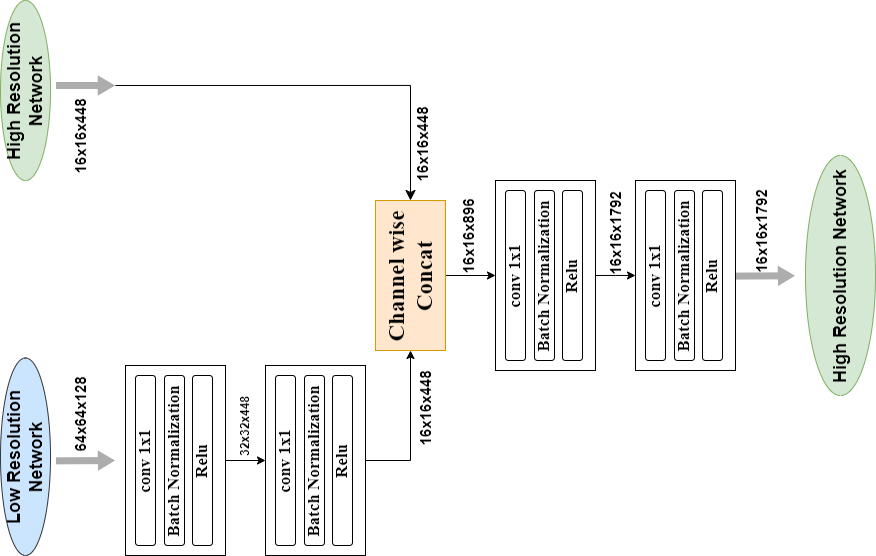}
    \caption{Features merge Logic connecting high resolution with low-resolution branch using $1 \times 1D$ convolution.}
    \label{fig:merge_logic}
\end{figure}

As referred to in Figure~\ref{fig:model}, the model that we have finalized is an extension of HookNet Architecture with multiple UNet in parallel which enhances the feature-capturing capability across resolutions. The low-resolution UNET is used for transferring global structure context to high-resolution UNET having local features. There are multiple ways we can combine multiple UNETs together. HookNet crops the high dimensions feature map to concatenate with the lower dimension features. Due to cropping, feature maps from the border regions are lost. We have implemented a custom merge logic that uses multiple 1D convolutions to non-linearly compress the high-dimension features to lower dimensions before concatenation as seen in Figure~\ref{fig:merge_logic}. This stepwise compression of features makes sure that we use full global features leading to better performance of the model for objects larger than FOV. As shown in Figure~\ref{fig:merge_logic}, we have also used multiple 1D convolutions to compress the depth of the feature vectors after merging, to effectively capture the local patterns and dependencies from multiple branches of UNet. Multiple 1D convolutions ensure non-linear feature compression during merging too.

Traditional papers generally use the last block of the UNet Encoder to merge the features of multiple UNets. Due to this, complex abstract features of high resolution get merged with low-resolution features. Our model uses the second last encoder block of higher resolution UNet for merging (refer to Figure~\ref{fig:model}). This helps the model to combine more generic low-level features with global context. To generate a better output mask at the last layer of the decoder, we have combined the output of both the UNet channel-wise by cropping the center of the low-resolution image, resizing it to high-resolution dimensions, and concatenating it with 1D convolutions. This helps the model gain more confidence in the predicted class of the pixel.

We have used EfficientNetB4~\cite{tan2019efficientnet} Architecture as the encoder for both U-Net. As stated by the author in the Efficient paper ~\cite{tan2019efficientnet}, the EfficientNet model is designed to achieve state-of-the-art performance with very little model size and trainable parameters. They achieve this efficiency by uniformly scaling width, depth \& resolutions in parallel. We have tried multiple Resolution levels like 0/2, 0/3, 1/4, 1/3. The best model results were given by resolution level 1/3. Our model currently generates segmentation for only a $40\times$ resolution level with 3-channel output. Each channel represents 1 class. We have also tried multiple pixel-aligned context transfers at various levels of decoders. But the gain from 1 connect to 3 connect between UNet was marginal. So, we have taken only 1 context transfer merge as shown in Figure~\ref{fig:merge_logic}.

We have used a patch size of $512 \times 512$ with 3-channel RGB input and $512 \times 512$ with 3-channel class output. Refer to Table~\ref{tab:hyperparameters_used} for hyperparameters used in training. We have used a stepwise cyclic learning rate starting from 0.001 with a decreased jump at each cycle. This helps models find the lowest point in global minima instead of getting stuck in local minima.

\begin{table}[!b]
\centering
\begin{tabular}{l|l}
\hline
\textbf{Parameter }         & \textbf{Values }                                                                                                                \\ \hline
Input              & {[}($512 \times 512 \times 3$), ($512 \times 512 \times 3$){]} \\ \hline
Output             & {[}($512 \times 512 \times 3$){]}                                                          \\ \hline
Optimizer          & Adam optimizer                                                                                                                \\ \hline
Initial Weights    & Random weight Initialization                                                                                           \\ \hline
Loss function      & Weighted {[}Categorical Cross   Entropy + Dice   Loss{]}                                                               \\ \hline
Learning Rate (LR) &     \begin{minipage}{.40\linewidth}
            \includegraphics[width=0.95\linewidth]{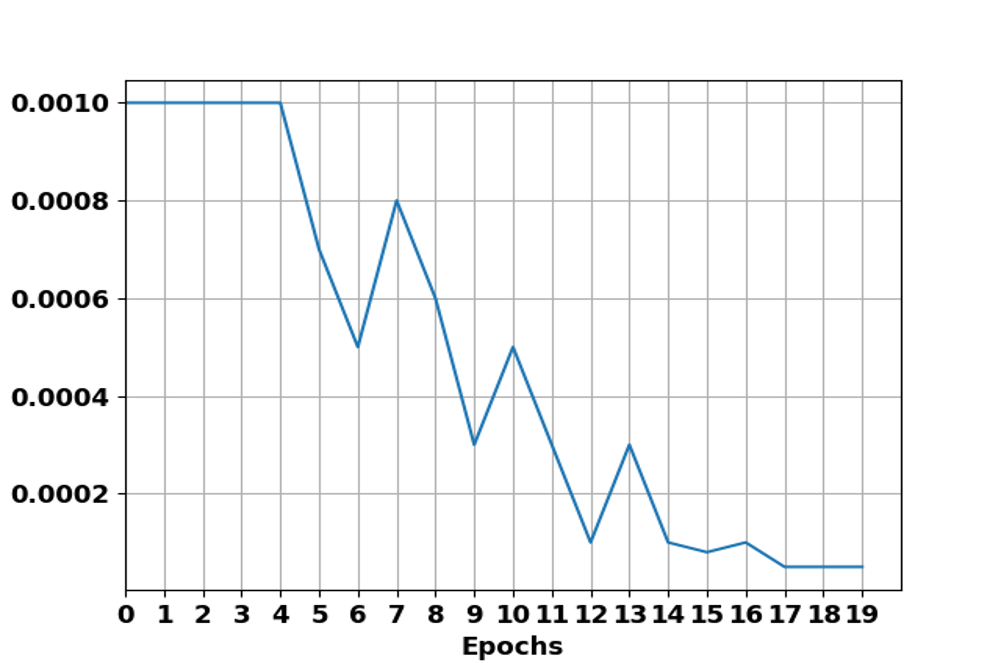}
            \end{minipage}                                                                                                                   \\ \hline
\end{tabular}
\caption{Hyperparameters that are used in the model's training.}
\label{tab:hyperparameters_used}
\end{table}

\subsection*{Loss Function}
Cross Entropy~\cite{zhang2018generalized} loss outputs probability estimates \& works well with multi-class classification tasks whereas dice loss~\cite{sudre2017generalised} is used for smooth and proper boundaries of the segmented regions. We have used the combination of Weighted Cross Entropy (WCE) loss with weighted dice loss (WDL) for all the pixel labels $Y$ and predicted label $\hat{Y}$ at all the output branches of UNet.

\begin{align}
\label{eq:ws}
     W^s &=  \begin{bmatrix}
        W_0^s & W_1^s & \cdots & W_L^s
    \end{bmatrix} \\
    \label{eq:wdc}
    W^D_c &= \begin{bmatrix}
        W_0^D & W_1^D & \cdots & W_L^D
    \end{bmatrix} \\
    \label{eq:wce}
    WCE &= \sum_{i=0}^{B}\sum_{j=0}^{C}\sum_{k=0}^{w\cdot h} {W_j^D}\cdot {Y_{ij}^{k}\cdot {\log (\hat{Y_{ij}^k})}} \\
    \label{eq:wdl}
    WDL &= \sum_{j=0}^{C} \left[ W_j^D \cdot \left[ \frac{2 \cdot \sum_{i = 0}^{w\cdot h} \left( Y_j^i \cap \hat{Y_j^i} \right)}{\sum_{i=0}^{w\cdot h} Y_j^i \cup  \sum_{i=0}^{w\cdot h} \hat{Y_j^i} + \epsilon} \right] \right] \\
    \label{eq:total_loss}
    Total~Loss &= W^s \cdot \begin{bmatrix} (WCE + WDL)_0 \\
    \vdots \\ 
    (WCE + WDL)_L \end{bmatrix} 
\end{align}

As shown in equation~\ref{eq:wce} \&~\ref{eq:wdl}, Both the losses are dynamically weighted (discussed in class imbalance
section) by multiplying $W^D$ to the losses across batch $B$. The results are then summed together for all the
classes $C$ to get combined loss. As we have multiple branch output, the dot product of combined loss
with the static weights $W^s$ gives \textit{total loss} for each resolution level $L$ referring in equation~\ref{eq:total_loss}.

\subsection*{Training}
\begin{figure}
    \centering
    \includegraphics[width=\linewidth]{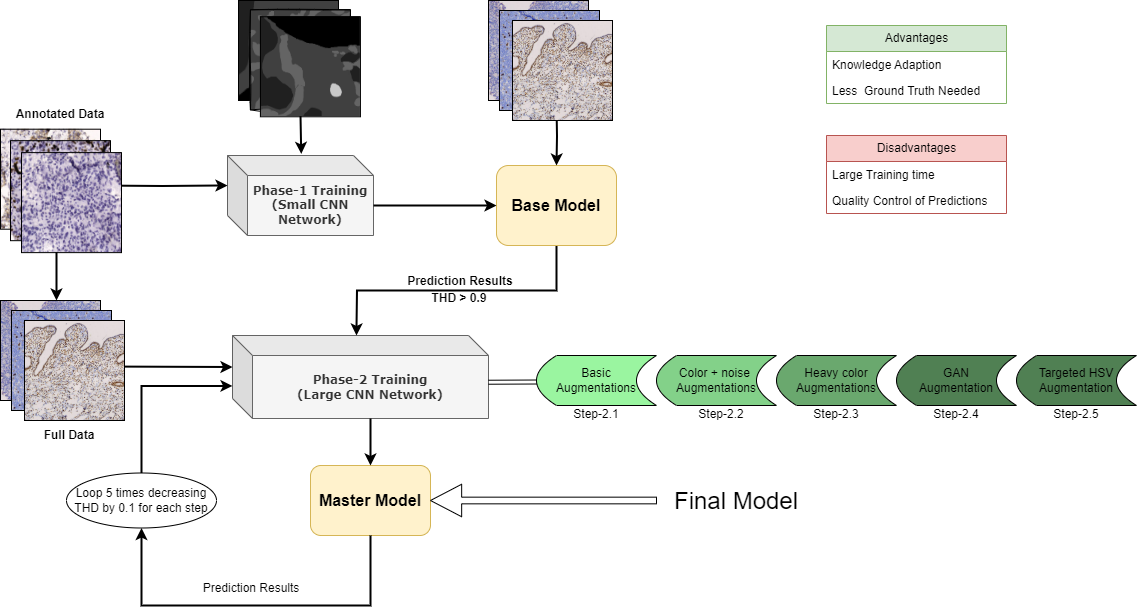}
    \caption{Multi-Step Training Process which starts with training base model and improving it by feeding its predictions with increased information and noise.}
    \label{fig:multi_step}
\end{figure}

Since we have only a few ground truth annotations with impurities in terms of the entire dataset, we have used the 2-phase training approach. In this approach, we train a small base model (EfficientNetB0) with given partial annotations. The Model is trained for 20 epochs performing the best it can on the given dataset. This model gave a decent performance of 0.70 IOU but is not generic enough for multiple scanners and data sources. This model is now used to predict the entire dataset available (both with and without annotations). The threshold was kept very high (0.9) to reduce the noise in the predictions. Now this predicted dataset is fed to a larger model (EfficientnetB4) for training with increased augmentations and decreased threshold. This process is repeated 5 times for best performance as referred to in Figure~\ref{fig:multi_step}. The idea here is that in every step the model will learn the parent model plus some additional learning on its own.

Due to the small manually annotated dataset, In Phase-1 model mostly learns the basic nuances such as cell structures, background, regions to ignore, different tissue types, staining colors, etc. However, this model is not able to scale to different data sources, scanner variations, and similar-looking tissue like DCIS, and acinis which look like tumors in a smaller field of view (FOV). In Phase 2, with the help of increasing augmentations and predictions, the model goes through a lot of variations of data. Hence model becomes increasingly capable of handling wider data distributions.

Table~\ref{tab:hyperparameters} describes the increasing augmentations and decreasing thresholds on each step. During base model training, the patch-level classes were appropriately balanced. So, we have used static class weights for base model training. Static class weights were also used in the last child training to make the model a little biased toward tumor regions. In the rest of the intermediary step, batch-level pixel-wise dynamic weights (referred to in the class imbalance section) were used because of the high pixel-level class imbalance. The base model was trained with considerably basic augmentations (refer augmentation section). During each master model training, we slightly kept on increasing the augmentation types/ranges such that the model learns more varied distributions as described in Table~\ref{tab:hyperparameters}. For details of each augmentation in Table~\ref{tab:hyperparameters} please refer to the augmentation section.

\begin{table}[!t]
\centering
\resizebox{\textwidth}{!}{%
\begin{tabular}{lccllccccl}
\hline
Step              & \multicolumn{1}{l}{Model} & \multicolumn{1}{l}{Encoder} & Dataset                                                                          & Augmentations                                                                                                                        & \multicolumn{1}{l}{Threshold} & \multicolumn{1}{l}{Epochs} & \multicolumn{1}{l}{LR0} & \multicolumn{1}{l}{Batch Size} & Class Weight \\ \hline
Base Training     & UNet                      & B0                          & \begin{tabular}[c]{@{}l@{}}Manually annotated \\ slides\end{tabular}             & basic - crop, flip, rotate, rescale, normalize                                                                                       & 0.9                           & 20                         & 0.01                    & \multirow{6}{*}{16}            & Static       \\ \cline{1-8} \cline{10-10} 
Master Training-1 & \multirow{5}{*}{m-UNet}   & \multirow{5}{*}{B4}         & \multirow{5}{*}{\begin{tabular}[c]{@{}l@{}}Complete training\\ set\end{tabular}} & \begin{tabular}[c]{@{}l@{}}basic + color jitter \\ + noise addition\end{tabular}                                                     & 0.8                           & 5                          & 0.005                   &                                & Dynamic      \\ \cline{1-1} \cline{5-8} \cline{10-10} 
Master Training-2 &                           &                             &                                                                                  & \begin{tabular}[c]{@{}l@{}}basic \\ + heavy color augmentation\\ + noise addition\end{tabular}                                       & 0.7                           & 5                          & 0.001                   &                                & Dynamic      \\ \cline{1-1} \cline{5-8} \cline{10-10} 
Master Training-3 &                           &                             &                                                                                  & \begin{tabular}[c]{@{}l@{}}Targeted HSV augmentation \\ + basic   \\ + heavy color  augmentation   \\ + noise addition\end{tabular}  & 0.6                           & 5                          & 0.0007                  &                                & Dynamic      \\ \cline{1-1} \cline{5-8} \cline{10-10} 
Master Training-4 &                           &                             &                                                                                  & \begin{tabular}[c]{@{}l@{}}GAN + targeted HSV augmentation\\ + basic  \\ + heavy color augmentation \\ + noise addition\end{tabular} & 0.5                           & 5                          & 0.0005                  &                                & Dynamic      \\ \cline{1-1} \cline{5-8} \cline{10-10} 
Master Training-5 &                           &                             &                                                                                  & \begin{tabular}[c]{@{}l@{}}GAN augmentation + XY coordinate Jitter \\ + basic   \\ + noise  addition\end{tabular}                    & 0.4                           & 10                         & 0.0001                  &                                & Static       \\ \hline
\end{tabular}%
}
\caption{Stepwise hyperparameters such as epochs, learning rate, class weights, etc. that are used for training.}
\label{tab:hyperparameters}
\end{table}
\bibliography{roi_paper_reference}

\section*{Acknowledgements}
We would like to thank the management at Applied Materials for supporting our work in this domain. We are deeply grateful to the pathologists who provided IHC scores annotations for this work, including Dr. Kanthilatha Pai, Dr. Brij Mohan Kumar Singh, Dr. Anil Betigeri, Dr. Vani Verma, and Dr. Madhavi Pai. We also like to extend the gratitude to all the domain experts involved in the knowledge sharing, slide quality control, and output validation.

\section*{Author contributions}
S.J., A.M., G.S., S.M., and R.K. were involved in the planning and designing of the experiments. S.J., and K.A., were involved in acquiring the data to perform the experiments and provide strategic support. S.J. and A.M. wrote the code to achieve different tasks. G.S. and S.M. were involved in validating the annotation of the data used during the training process. The results collection and performance analysis was conducted by A.M., and R.K.. S.J., A.M., and P.M. wrote the manuscripts with the assistance and feedback of all other co-authors.

\section*{Competing interests}
The author declares no competing interests.

\section*{Data Availability}
The datasets used in this study were collected as per an internal agreement between Applied Materials India Pvt. Ltd. and hospitals/laboratories participating in this study. The slides and doctor's annotations are not available publicly due to restrictions in the data-sharing agreements with the participating institutions. Data are however available from the authors upon reasonable requests and with the permission of competent authority at Applied Materials India Pvt. Ltd. and participating institutions.




\end{document}